\documentclass[lettersize,journal]{IEEEtran}
\usepackage{amsmath,amsfonts}
\usepackage{array}
\usepackage[caption=false,font=normalsize,labelfont=sf,textfont=sf]{subfig}
\usepackage{textcomp}
\usepackage{stfloats}
\usepackage{url}
\usepackage{verbatim}
\usepackage{multirow}
\usepackage{graphicx}
\usepackage{cite}
\usepackage{makecell}
\usepackage{booktabs}
\usepackage{bm}
\usepackage[lined,boxed,linesnumbered,ruled]{algorithm2e}
\begin{document}

\title{A Risk-aware Planning Framework of UGVs in Off-Road Environment}

\author{Junkai~Jiang$^{1}$, Zhenhua~Hu$^{2}$, Zihan~Xie$^{2}$, Changlong~Hao$^{2}$, Hongyu~Liu$^{2}$, Wenliang~Xu$^{2}$, Yuning~Wang$^{1}$, Lei~He$^{1\ast}$, Shaobing~Xu$^{1\ast}$, and Jianqiang~Wang$^{1\ast}$
\thanks{This research was funded by National Natural Science Foundation of China, Science Fund for Creative Research Groups (Grant No. 52221005) and National Natural Science Foundation of China (Grant No. 52131201). This research was also supported by the Tsinghua University-Didi Joint Research Center for Future Mobility.}
\thanks{$^{1}$Junkai~Jiang, Yuning~Wang, Lei~He, Shaobing~Xu, Jianqiang~Wang are with the School of Vehicle and Mobility, Tsinghua University, Beijing, China.}
\thanks{$^{2}$ Zhenhua~Hu, Zihan~Xie, Changlong~Hao, Hongyu~Liu, Wenliang~Xu are interns with the Lab of Intelligent and Connected Vehicles, Tsinghua University, Beijing, China.}
\thanks{$^{\ast}$Corresponding author: Lei~He (helei2023@tsinghua.edu.cn), Shaobing~Xu (shaobxu@tsinghua.edu.cn) and Jianqiang~Wang (wjqlws@tsinghua.edu.cn).}
}



\maketitle

\begin{abstract} 
Planning module is an essential component of intelligent vehicle study. In this paper, we address the risk-aware planning problem of UGVs through a global-local planning framework which seamlessly integrates risk assessment methods. In particular, a global planning algorithm named Coarse2fine A* is proposed, which incorporates a potential field approach to enhance the safety of the planning results while ensuring the efficiency of the algorithm. A deterministic sampling method for local planning is leveraged and modified to suit off-road environment. It also integrates a risk assessment model to emphasize the avoidance of local risks. The performance of the algorithm is demonstrated through simulation experiments by comparing it with baseline algorithms, where the results of Coarse2fine A* are shown to be approximately 30\% safer than those of the baseline algorithms. The practicality and effectiveness of the proposed planning framework are validated by deploying it on a real-world system consisting of a control center and a practical UGV platform.
\end{abstract}



\section{Introduction} \label{Sec:Introduction}
\IEEEPARstart{R}{esearch} on Unmanned Ground Vehicles (UGVs) in off-road environment has attracted growing attention in recent years, for their capability of accomplishing dangerous tasks independently, thus reducing the risks human beings may face. UGVs have extensive applications, such as disaster relief \cite{mandow2020experimental}, hazard detection \cite{tan2020unmanned}, target destruction \cite{nohel2020combat}, etc. To equip UGVs with ability of autonomous driving, the integration of technologies such as localization, perception, planning, and control is required \cite{yang2019practical}. Among these technologies, planning module is one of the core components. Reliable path and trajectory planning are indispensable to ensure UGVs executing tasks successfully in off-road environment.

The planning module aims to generate a safe and efficient trajectory for UGVs, based on the results of perception, localization, and destination-related information. Planning is generally categorized into global and local planning. Global planning, akin to navigation, involves creating a path from the current position to the destination using a relatively static global map \cite{liu2017global}. It is typically executed only once and is re-planned when necessary. Local planning, on the other hand, takes the global planning results as a reference line and iteratively plans the trajectory for UGVs to follow. It considers dynamic obstacles and real-time sensor data, with the planned trajectory subsequently transmitted to the control module. This paper focuses on both global and local planning for UGVs.

Compared to urban traffic environment, the off-road environment exhibits a distinctly higher potential for risks that UGVs may encounter. Firstly, the terrain in off-road settings is markedly more complex, characterized by the scarcity of well-maintained roads and the presence of challenging surfaces such as sand, gravel, and grass. Secondly, the lack of traffic regulations presents a significant challenge in predicting the intentions of other entities within the environment, thereby intensifying the overall risk situation. Consequently, it is essential to develop risk-aware planning methods in off-road environment.

Currently, research on planning amidst various risks tends to focus on specific domains, such as terrain conditions \cite{moore2023ura}, multi-vehicle conflicts \cite{wang2023risk}, or the uncertainty of intentions \cite{aoude2013probabilistically}. Our purpose is to adopt a unified approach to comprehensively describe various risks and provide support for the planning module. To this end, we focus on a risk-aware planning framework for UGVs in off-road environments, emphasizing comprehensive risk assessment for such terrains. The objective is to integrate risk assessment methods with both global and local planning modules, incorporating them into an overall planning framework. We are committed to validating the effectiveness of the proposed framework through practical experiments.

\begin{figure}
    \centering
    \includegraphics[width=0.8\linewidth]{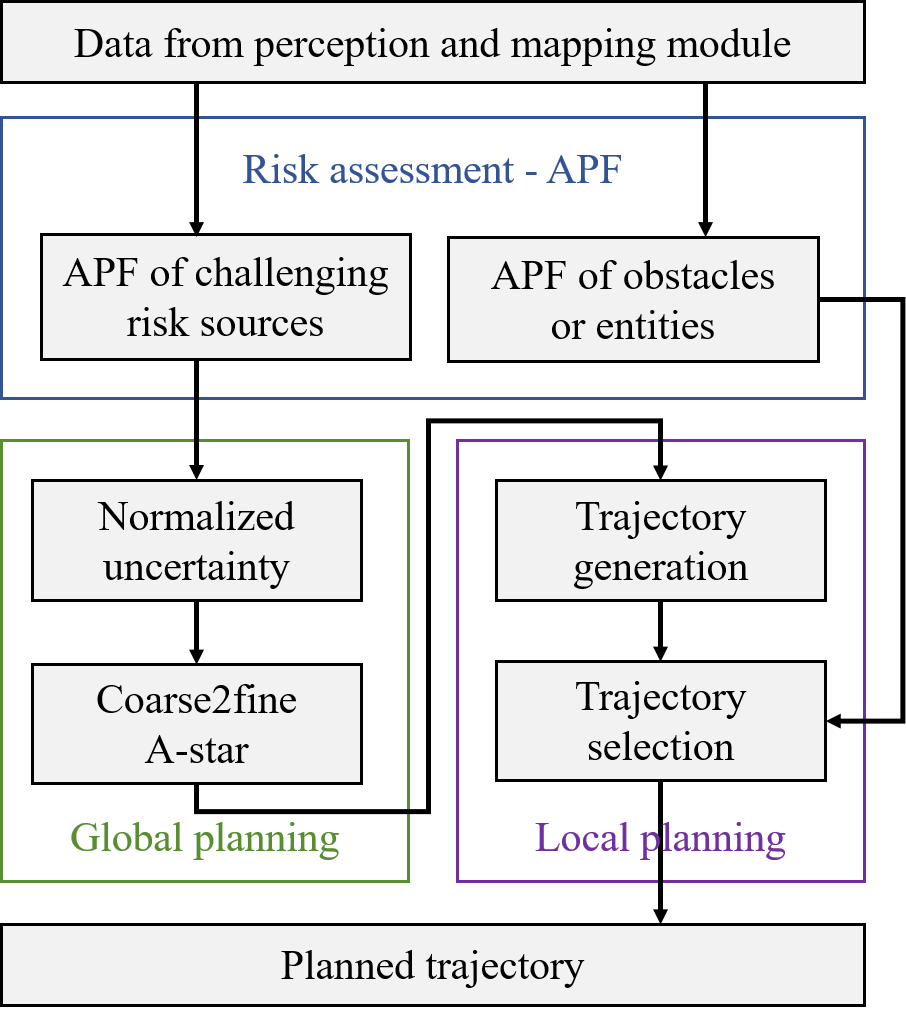}
    \caption{The framework of proposed system.}
    \label{fig1}
\end{figure}

In further detail, we present a novel framework employing the concept of artificial potential field (APF) to characterize diverse risks and incorporate the outcomes of risk assessment into the planning module of UGVs. The global planning component has been improved, building upon our previous work \cite{jiang2023global}, to enhance both its efficiency and flexibility. For local planning, a deterministic sampling approach has been employed to balance the quality and efficiency of the planning results. The system framework is illustrated in Fig. \ref{fig1}.

The contributions of this paper are as follows:
\begin{enumerate}
    \item {a planning framework of UGVs that seamlessly integrates risk assessment methods,}
    \item {a global planning algorithm named Coarse2fine A*, which takes risk into account and enhances both efficiency and flexibility,}
    \item {a refined deterministic sampling method suitable for off-road environments, which underlines the awareness of local risks,}
    \item {the deployment of the proposed method on a practical UGV, validating the effectiveness of the framework.}
\end{enumerate}

The remainder of this paper is organized as follows. Section \ref{Sec:RelatedWork} provides a concise review of related work, while section \ref{Sec:SystemFramework} introduces the problem description and our proposed framework. Subsequently, Sections \ref{Sec:RiskAssessment} through \ref{Sec:Local Planning} offer a comprehensive exposition on risk assessment, global planning, and local planning, respectively. The results of simulations and practical experiments are detailed in section \ref{Sec:Simu&Expe} to demonstrate the performance and practicality of our framework. Finally, section \ref{Sec:Conclusion} concludes this paper and provides directions for future work.

\section{Related Work} \label{Sec:RelatedWork}
Risk-aware planning for UGVs entails the evaluation of environmental risks, identification of reference path with lower risk level, and generation of local trajectory with an emphasis on safety. Unlike certain studies that integrate obstacle avoidance directly into the cost function of planning \cite{hang2020integrated,khaitan2021safe,ren2020optimal}, we treat risk assessment as an independent module to underscore the safety of UGVs. Several reviews have summarized state-of-the-art motion planning techniques in area of autonomous driving, including pipeline method and end-to-end method \cite{teng2023motion,katrakazas2015real,schwarting2018planning}. The latter, however, has yet demonstrated sufficient stability and interpretability. Therefore, this paper adopts the pipeline planning method. In this section, we explore related work in the fields of risk assessment, global planning and local planning.  

\subsection{Related Work in Risk Assessment}
In off-road environment, complex terrains represent a significant source of risk, thereby making terrain traversability analysis a key area of interest in off-road risk assessment \cite{borges2022survey}. There is also a growing interest in merging traversability analysis with motion planning to generate paths or trajectories that are aware of risks \cite{moore2023ura,cai2023probabilistic}. Elevation differential is another risky factor. In \cite{liu2017global}, an improved A* algorithm was proposed, further considering the elevation of adjacent nodes to enhance the safety of the generated path. Additionally, various studies have aimed to integrate different risks and produce output in the form of risk maps or uncertainty maps \cite{wang2023aerial}, which are then applied to the planning and control modules of UGVs. In recent years, the potential field approach has been explored for modeling diverse risk types within off-road environments \cite{hu2021integrated,tian2023driving}. However, special attention is required to ensure the validity and rationality of such models.

In traffic environment, risk assessment pays more attention on the interactive influences among various entities. For instance, in longitudinal risk assessment methods for vehicles, metrics such as Time Headway (THW), Time-to-Collision (TTC), and their derivatives are commonly used as safety indicators \cite{kondoh2008identification}. Nonetheless, these metrics are context-specific and might not offer a comprehensive assessment across varying environments. To overcome this limitation, the APF method has been thoroughly investigated. In \cite{ni2011unified}, a vehicle motion model based on potential field theory was developed, facilitating precise micro-level risk assessment within traffic flow. Furthermore, the concept of driving safety field is proposed in \cite{wang2015driving,wang2016driving}, establishing a unified risk assessment model for the human-vehicle-road environment and enabling systematic evaluation of risks in traffic environment.

In summary, APF possesses the capacity to integrate all potential risk factors into a unified framework for assessment. As a consequence, this paper will adopt APF as the risk assessment method to ensure safety.

\subsection{Related Work in Global Planning}
Considerable efforts have been invested in exploring global path planning for UGVs, with the algorithms broadly falling into two categories: graph search and space sampling.

For graph search methods, the primary idea is to sequentially generate crucial path nodes within a certain space, followed by identifying a feasible path between the current node and the destination. Notable algorithms in this category include the Visibility Graph \cite{lozano1979algorithm}, Voronoi Graph \cite{aurenhammer1991voronoi}, Dijkstra search \cite{dijkstra2022note}, A* search \cite{hart1968formal}, and their variations. The A* search, in particular, combines the advantages of Dijkstra search and greedy algorithm, offering theoretical guarantees of solution optimality. However, the real-time performance of A* algorithm is often considered inadequate. While we have implemented efficiency improvement methods in our previous work \cite{jiang2023global} to reduce time consumption, this strategy still lacks the desired flexibility and leaves space for further refinement.

Regarding space sampling methods, the basic idea is to connect the current node and the destination by random or deterministic sampling. Representative methods include Probability Roadmap Method (PRM) and Rapidly-exploring Random Tree (RRT). PRM is suitable for path planning problems characterized by high dimensionality and multiple constraints \cite{kavraki1996probabilistic}. However, it carries a risk of planning failure, particularly when the sampling density is insufficient. Moreover, the use of probabilistic sampling may miss critical nodes in the spatial environment, potentially compromising path optimality.  RRT is a widely-used path planning method, which randomly samples nodes in the configuration space and rapidly explores the tree structure to efficiently discover feasible paths \cite{cheng2002resolution}. Although RRT can quickly produce feasible paths, the inherent randomness of the algorithm often results in suboptimal solutions. Despite the existence of numerous variations of the PRM and RRT algorithms \cite{lin2006gaussian,ravankar2020hpprm,gammell2014informed,chi2018risk}, the core challenges associated with these methods remain unaddressed.

In this paper, we propose a global planning algorithm named Coarse2fine A* (building upon \cite{jiang2023global}) to enhance the flexibility and efficiency, and integrate it with the risk assessment module.

\subsection{Related Work in Local Planning}
Local planning aims to generate a safe and efficient local trajectory on the basis of global path. It is worth noting that the output of local planning must be a trajectory, which includes both path and speed profiles \cite{huang2019motion}. Given that planning results are inherently local and temporary, it is essential for local planning to operate in a receding-horizon manner to promptly adapt to changes in the environment. In \cite{teng2023motion}, local planning methods are segmented into three categories: state grid identification, primitive generation, and the combination of the two.

State grid identification can be subdivided into search-based (e.g., dynamic programming \cite{xu2014motion}), selection-based (e.g., greedy selection \cite{li2015real} and Markov decision process \cite{bai2015intention}), and optimization-based methods (e.g., nonlinear programming \cite{li2021optimization} and metaheuristics \cite{rosmann2017integrated}). Primitive generation method can also be further classified, including closed-form rules \cite{reeds1990optimal}, simulation-based \cite{dolgov2010path}, interpolation-based \cite{botros2022tunable,van2021cooperative}, and optimization-based methods \cite{hu2022adaptive}. The combination of the aforementioned methods can, in turn, give rise to a variety of trajectory construction approaches. 

Essentially, the goal of local planning is to solve an optimal control problem (OCP), aiming to minimize a pre-designed cost function subject to specific constraints. To achieve real-time performance, sampling strategies are often leveraged to solve the OCP \cite{werling2012optimal,xu2021system}. The basic idea is to generate a finite set of trajectories and subsequently identify the optimal one among them. \cite{xu2021system} introduced a deterministic sampling algorithm that decouples lateral and longitudinal motions within the Frenet coordinate system, samples across spatiotemporal dimensions to generate a set of candidate trajectories, and utilizes a carefully defined cost function for optimal selection. This paper adopts the deterministic sampling strategy and modifies the approach proposed in \cite{xu2021system} to better suit off-road environment while prioritizing safety.

\section{System Framework} \label{Sec:SystemFramework}
As illustrated in Fig. \ref{fig1}, the entire risk-aware planning system encompasses three primary components: the risk assessment module utilizing APF, the global planning module featuring the Coarse2fine A* algorithm, and the local planning module based on a deterministic sampling strategy. Indeed, achieving autonomous driving in off-road environment requires additional modules, such as perception, mapping, and control. This paper focuses on risk assessment and planning, with other modules being beyond its scope. We presuppose the availability of a basic topographical map of the current environment (for instance, obtained from open-source maps) and accurate information of obstacles, including position and velocity.

The risk assessment module is responsible for evaluating both global and local risks in the environment, drawing on data from the perception and mapping modules. The APF method is employed to assess static risk sources, such as non-accessible areas, off-road terrains, and risk centers, as well as the risks associated with dynamic obstacles or entities. This evaluation is conducted within a unified framework, with the information of the potential field being conveyed to subsequent modules.

The global planning module generates a global path, also known as a reference path, connecting the origin to the given destination, comprising a sequence of waypoints. In global planning, only the potential field generated by static risk sources is considered, while dynamic obstacles are addressed in local planning. Normalization of the static potential at each location on the global map is performed to produce an uncertainty map, which then feeds into the Coarse2fine A* algorithm, realizing efficient and safe global path planning.

Upon obtaining the reference path, the local planning module determines the near-optimal trajectory for UGVs to track. It first employs a sampling strategy to generate a number of terminal states around a future target point (4s in this paper). These terminal states are interconnected to the current state via a 5th-order polynomial, producing a set of candidate trajectories. From this set, feasible trajectories are identified considering vehicle dynamics and obstacle avoidance constraints. We further leverage a cost function to evaluate all feasible trajectories and select the optimal one. The potential field of obstacles and entities is considered in both the constraints checking and cost evaluation stages to enhance the safety of the local trajectory.

In the following three sections, the design of each module will be described in details.

\section{Risk Assessment} \label{Sec:RiskAssessment}
The off-road environment is composed of various elements such as rough terrains, uneven surfaces, natural obstacles, and potentially dangerous risk centers. Generally, risk elements in off-road environments can be classified into two main types. The first is static risk sources, such as non-accessible areas (e.g., mountains, buildings, forests, water bodies), off-road terrains (such as sand, grass, gravel), and risk centers (e.g., seismic sources, toxic substance leakage sites). These static risk sources are relatively stable, often extend over large areas, and should be avoided by UGVs as much as possible. They are suitable for consideration within the global planning module. The second type encompasses obstacles or entities, including other UGVs, rocks, trees, vegetation, and wildlife. These elements typically impact localized areas and may possess mobility, thereby exerting a less significant influence on the overall navigation strategy.  UGVs employ sensors (such as LiDAR and cameras) to detect and understand their surroundings and, during the local planning stage, select a trajectory that maximizes safety by avoiding collisions with these risk sources. The following involves establishing APF models for both categories of risk sources separately.

\subsection{APF of Static Risk Sources}
Here, a concise and unified APF modeling method \cite{moreau2017path} is adopted to quantify the risk levels associated with static risk sources. The potential field generated by them is defined as:
\begin{equation}
\begin{aligned}
    \label{eqn1}
    {E}_{S} = \begin{cases}
        0, \quad & r > r_{\max} \\
        k_S \cdot \cfrac{r^{n-2}-r_{\max}^{n-2}}{r_{\min}^{n-2}-r_{\max}^{n-2}}, \quad & r_{\min} < r < r_{\max} \\
        k_S, \quad & r < r_{\min} \\
    \end{cases}.
\end{aligned}
\end{equation}
where $E_S$ represents the potential field generated by a static risk source, and $r$ denotes the distance to it. The term $r_{\min}$ is the affirmative range around the risk source within which its influence is deemed significant, while $r_{\max}$ indicates the maximum extent of its influence. Within the range of $r_{\min}$, the potential field's intensity is determined by a coefficient $k_S$, which varies according to the risk source's nature. The more dangerous the source, the higher the value of $k_S$ (for example, a mountain would have a higher $k_S$ compared to grassland). Between $r_{\min}$ and $r_{\max}$, the potential field intensity gradually decreases. The variable n is the fractional order, which is set to 4 in this paper.  If a position falls within the influence range of multiple risk sources, the resultant potential field there is the sum of the individual influences generated by each of these risk sources. Thus, as a UGV approaches a risk source, the potential field generates repulsion, encouraging it to choose a safer path.

\begin{figure}[htbp]
  \centering
  \begin{minipage}{0.48\linewidth}
    \centering
    \includegraphics[width=0.85\linewidth]{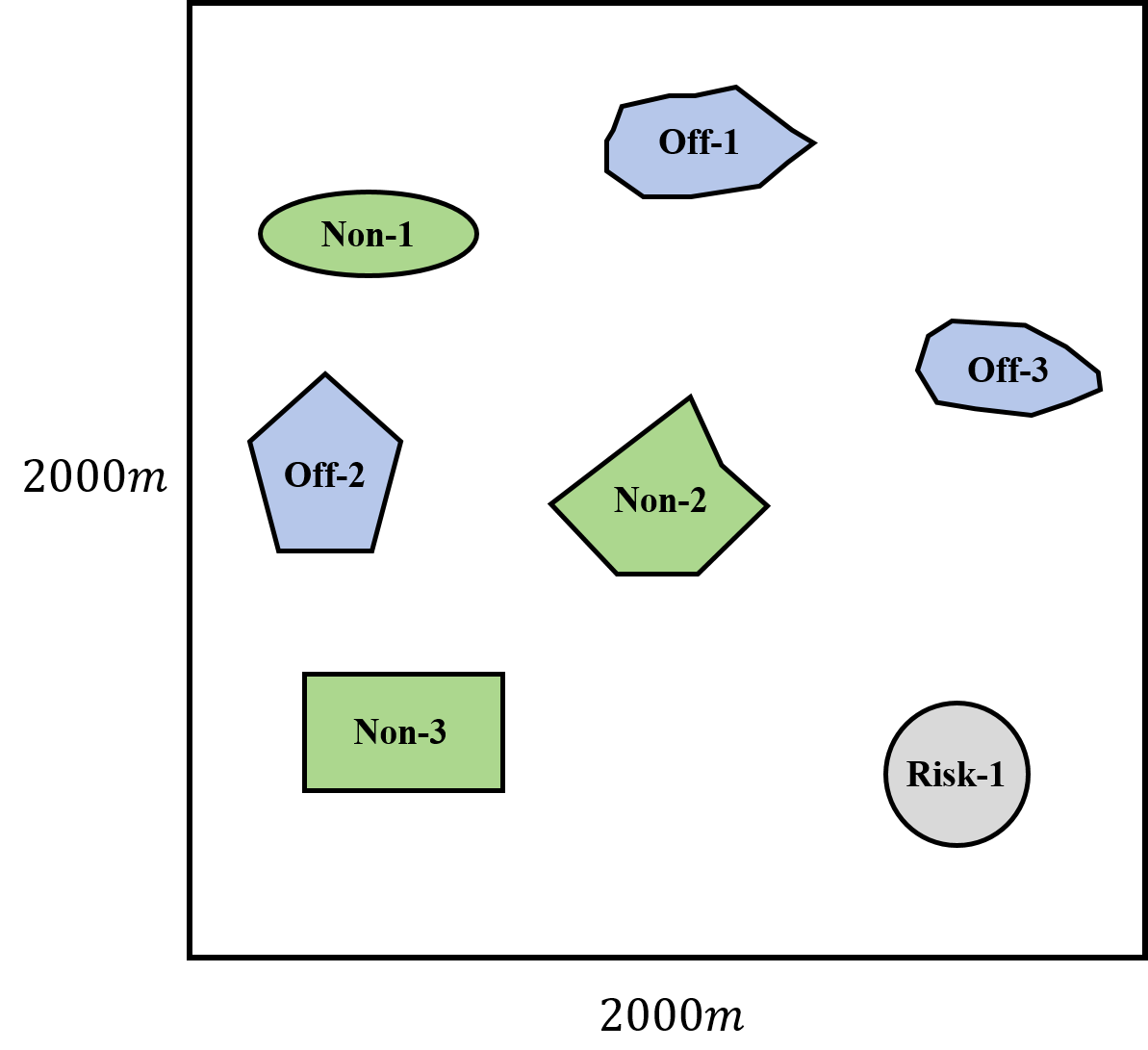} 
    \vspace{-2mm}
    \centerline{(a)}
  \end{minipage}
  \begin{minipage}{0.48\linewidth}
    \centering
    \includegraphics[width=1\linewidth]{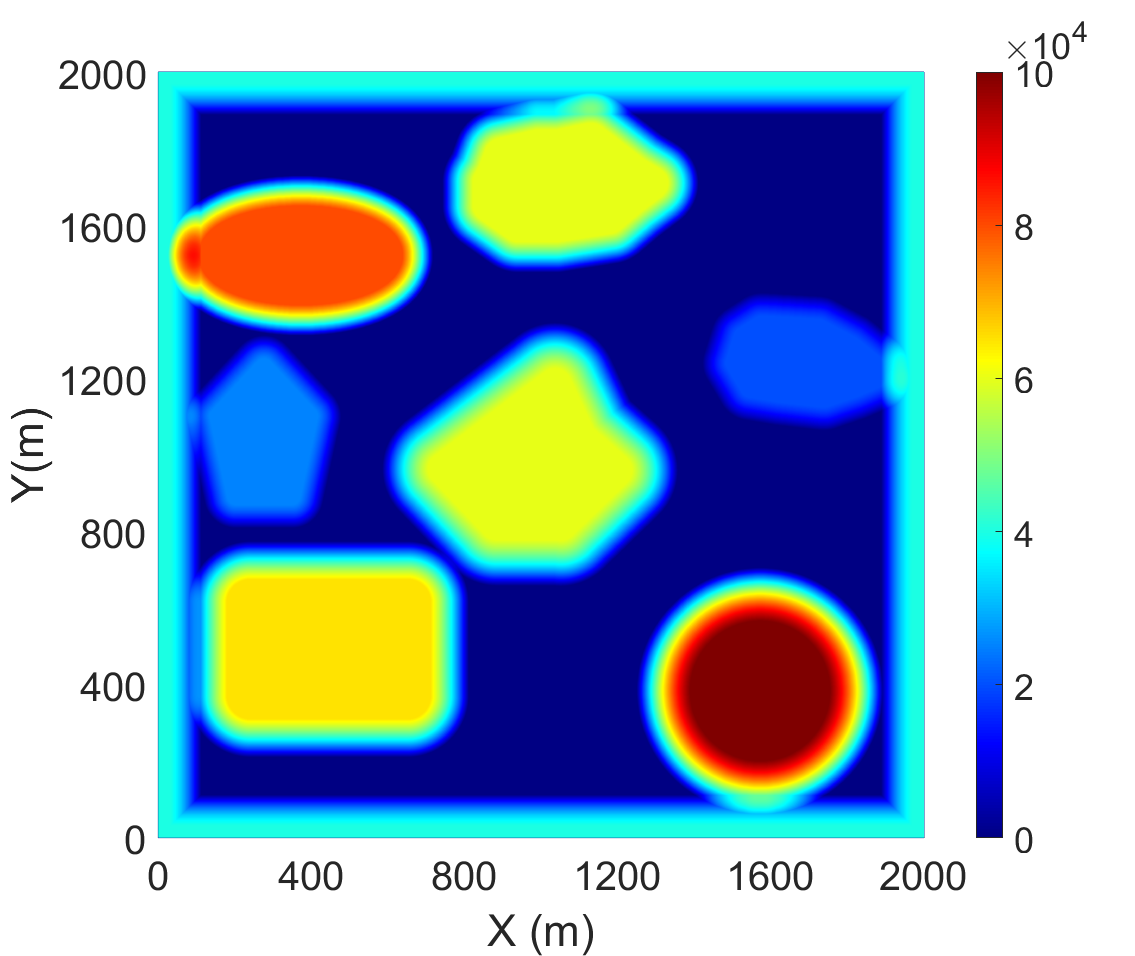} 
    \centerline{(b)}
  \end{minipage}
  \begin{minipage}{\linewidth} 
    \centering 
    \includegraphics[width=\linewidth]{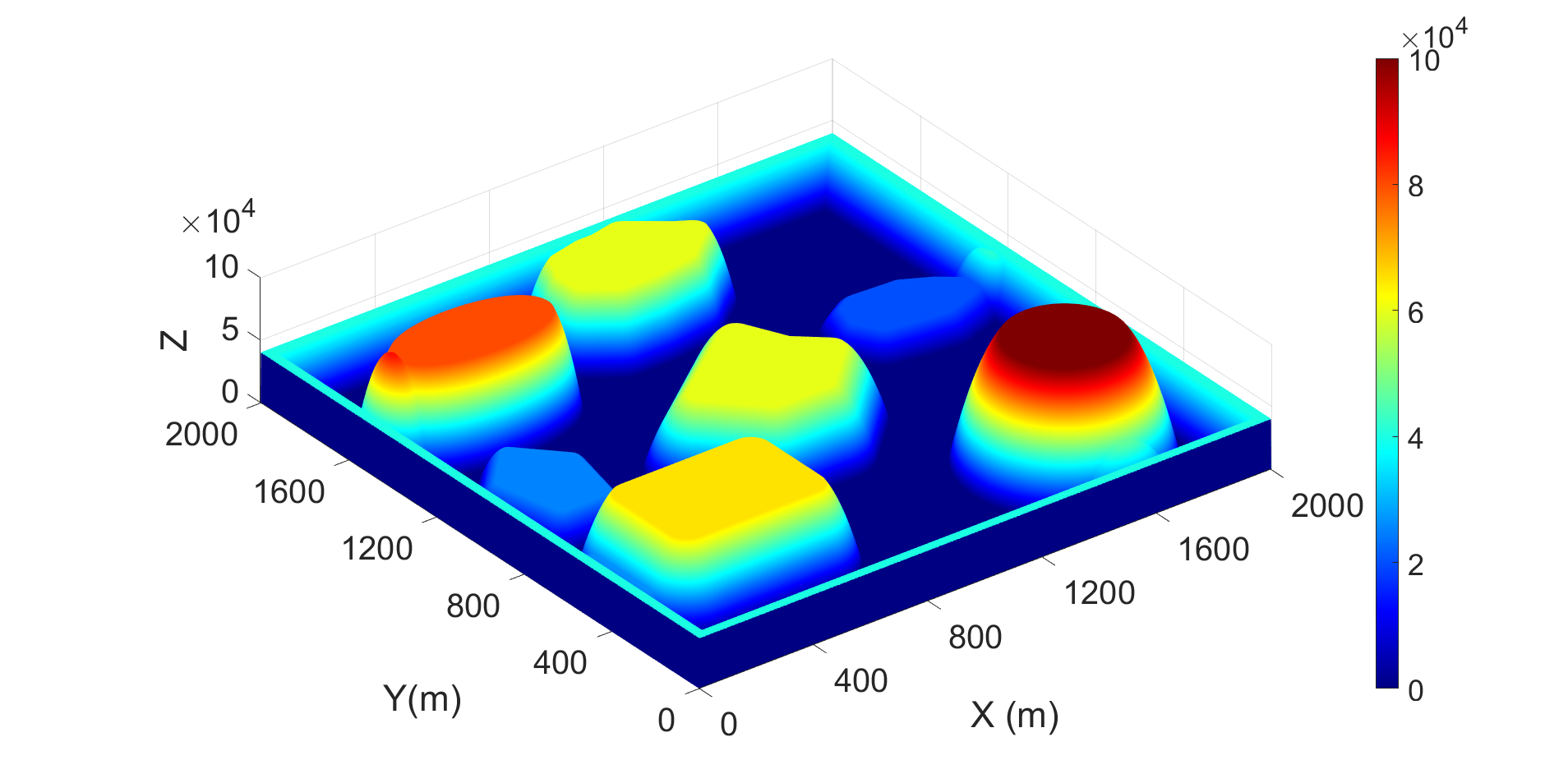} 
    \centerline{(c)}
  \end{minipage}
  \caption{The potential field map for static risk sources.}
  \label{fig2}
\end{figure}

Upon acquisition of map data, the potential field map for the off-road environment can be generated using the above-mentioned method. As depicted in Fig. \ref{fig2}(a), a 2000m×2000m region containing various non-accessible areas, off-road terrains, and risk centers is designed. Fig. \ref{fig2}(b) and Fig. \ref{fig2}(c) illustrate the 2D and 3D representations of the potential field maps for this scenario, where intensifying red indicates higher level of danger.

\subsection{APF of Obstacles or Entities}
In this part, the entities under consideration may possess mobility. Consequently, besides addressing static obstacles, the impact of their movement should also be considered to enhance safety. Our approach is inspired by the driving safety field model proposed in \cite{wang2015driving}, which is adapted to better fit off-road conditions. Fundamentally, the potential field of objects comprises both static and dynamic components, namely:
\begin{equation}
\label{eqn2}
    {E}_{OBS} = {\omega}_P E_P + {\omega}_D E_D
\end{equation}

For the static component $E_P$, it primarily relates to the object's inherent characteristics and its proximity:
\begin{equation}
\begin{aligned}
    \label{eqn3}
    {E}_{P} = \begin{cases}
        0, \quad & r > r_{\max} \\
        k \cdot r_P \cdot (\cfrac{1}{r^2} - \cfrac{1}{r_{\max}^2}), \quad & r_{\min} < r < r_{\max} \\
        E_{P\max}, \quad & r < r_{\min} \\
    \end{cases}.
\end{aligned}
\end{equation}
where $K$ represents the coefficient associated with the object itself (higher for more dangerous objects), $r_{\min}$ and $r_{\max}$ denote the collision distance and the maximum influence distance respectively, and $r_P$ is the distance coefficient, calculated as $r_P=r_{\min}^2 r_{\max}^2/(r_{\max}^2-r_{\min}^2)$, ensuring $E_P$ reach its peak when r is less than $r_{\min}$.

For the dynamic component $E_D$, it is additionally influenced by the object's velocity. An elevated velocity indicates a higher probability of potential risks ahead. The expression for $E_D$ is given by:
\begin{equation}
\label{eqn4}
    {E}_{D} = \frac{K}{r^{k_1}} \cdot \exp(k_2 \cdot v \cdot \cos(\theta))
\end{equation}
where $k_1$ and $k_2$ are the distance and velocity coefficients, v represents the object's velocity, and $\theta$ is the angle between vectors $\bm{v}$ and $\bm{r}$. The dynamic component $E_D$ similarly achieves a maximum value when the risks induced by the object's motion reach a certain threshold. 

\begin{figure}[htbp]
  \centering
  \begin{minipage}{0.48\linewidth}
    \centering
    \includegraphics[width=0.9\linewidth]{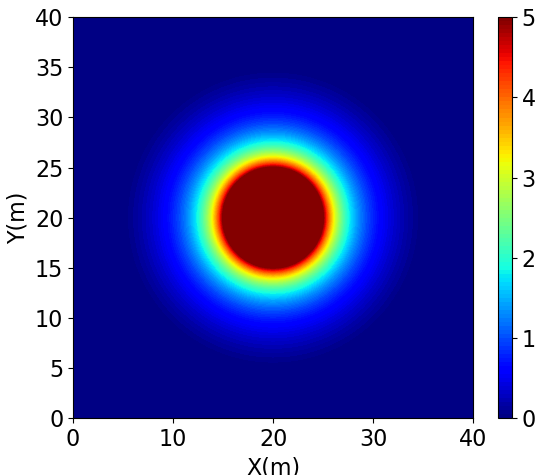} 
  \end{minipage}
  \begin{minipage}{0.48\linewidth}
    \centering
    \includegraphics[width=1\linewidth]{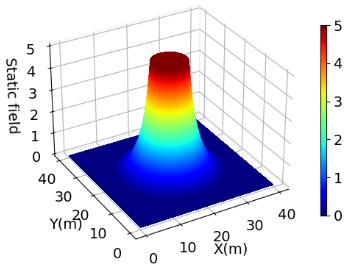} 
  \end{minipage}
  \centerline{(a) Static field}
  \begin{minipage}{0.48\linewidth}
    \centering
    \includegraphics[width=0.9\linewidth]{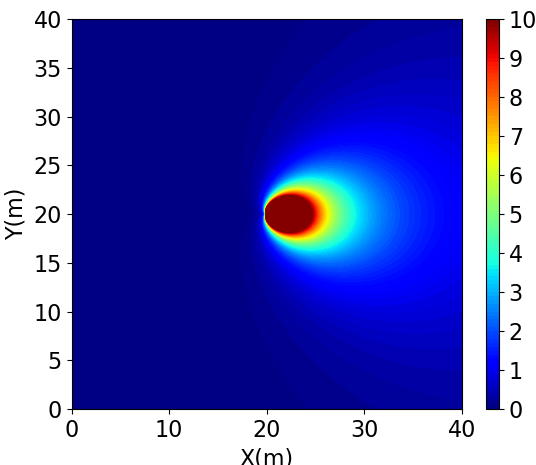}
  \end{minipage}
  \begin{minipage}{0.48\linewidth}
    \centering
    \includegraphics[width=1\linewidth]{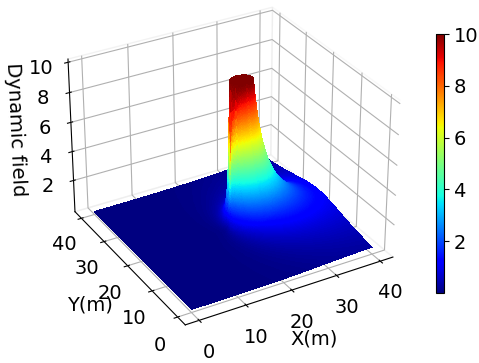} 
  \end{minipage}
  \centerline{(b) Dynamic field}
  \begin{minipage}{0.48\linewidth}
    \centering
    \includegraphics[width=0.9\linewidth]{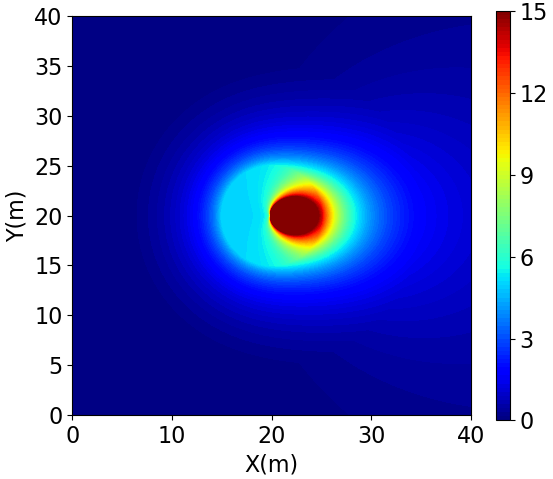} 
  \end{minipage}
  \begin{minipage}{0.48\linewidth}
    \centering
    \includegraphics[width=1\linewidth]{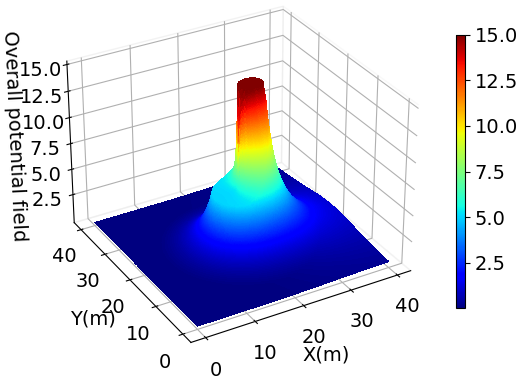} 
  \end{minipage}
  \centerline{(c) Overall potential field}
  \caption{The static field, dynamic field, and overall potential field of moving objects within a local area.}
  \label{fig3}
\end{figure}

The characteristics of objects, such as their type, size, position, and velocity, are acquired through the perception module. Fig. \ref{fig3} illustrates the static field, dynamic field, and overall potential field of moving objects within a local area. 

\section{Global Planning} \label{Sec:GlobalPlanning}
Through the risk assessment module presented in section \ref{Sec:RiskAssessment}, a comprehensive understanding of global risk situation and local environmental information is obtained. This section will explain the process of creating an uncertainty map based on the global risk context. Subsequently, a global path is devised utilizing the Coarse2Fine A* algorithm, which is then refined using Quadratic Programming (QP) for smoothness.

\subsection{Uncertainty Map Generation and Pre-processing}
\begin{figure}[htbp]
  \centering
  \begin{minipage}{0.48\linewidth}
    \centering
    \includegraphics[width=1\linewidth]{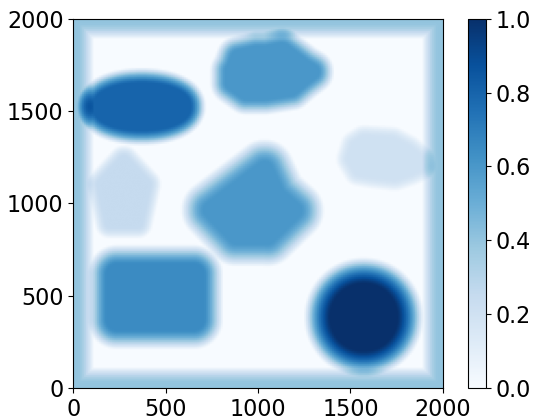} 
    \centerline{(a)}
  \end{minipage}
  \begin{minipage}{0.48\linewidth}
    \centering
    \includegraphics[width=1\linewidth]{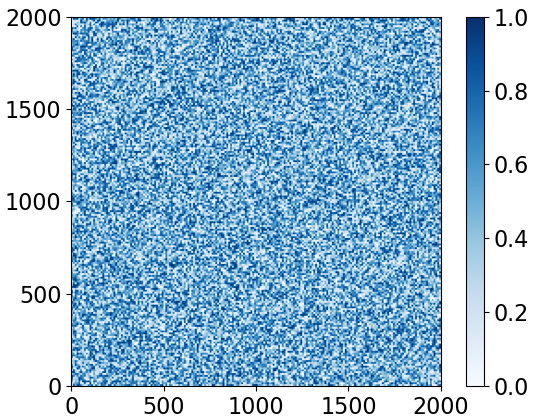} 
    \centerline{(b)}
  \end{minipage}
  \begin{minipage}{0.48\linewidth}
    \centering
    \includegraphics[width=1\linewidth]{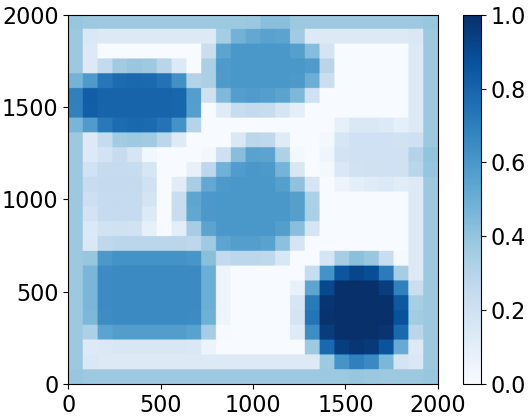} 
    \centerline{(c)}
  \end{minipage}
  \begin{minipage}{0.48\linewidth}
    \centering
    \includegraphics[width=1\linewidth]{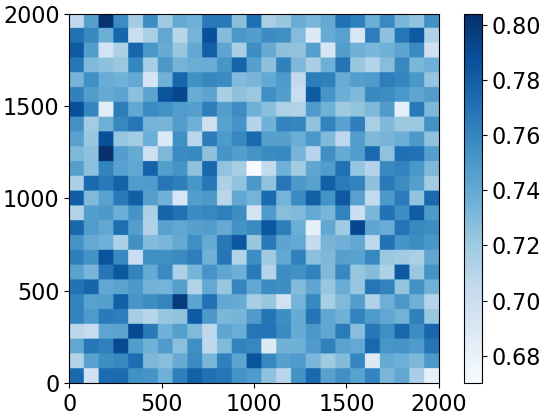} 
    \centerline{(d)}
  \end{minipage}
  \caption{The uncertainty maps: (a) and (c) are fine map and coarse map of the designed scenario; (b) and (d) are fine map and coarse map with randomly generated uncertainty.}
  \label{fig4}
\end{figure}

As defined in \cite{jiang2023global}, areas with higher potential field in off-road environment are considered with high area uncertainty, while areas with lower potential field are the opposite. If the potential field at a specific location surpasses a predetermined threshold, it is regarded as a non-accessible area in path planning. Normalizing the potential field values across different locations in Fig. \ref{fig2} yields the corresponding uncertainty map for the designed scenario, as illustrated in Fig. \ref{fig4}(a). Alternatively, as shown in Fig. \ref{fig4}(b), we can also obtain a map of area uncertainty randomly generated across various locations to compare algorithm performance in simulation experiments.

Within the Coarse2fine A* algorithm, it is necessary to process the original uncertainty map into both fine and coarse representations. The grid resolutions in Fig. \ref{fig4}(a) and \ref{fig4}(b) stand at $10m$, denoting the fine map. The derivation of the coarse map is achieved via a mixed pooling method, defined as follows:
\begin{equation}
\label{eqn5}
    M_{ij}^{\textnormal{coar}} = \lambda \cdot \max_{(p,q) \in \mathcal{R}_{ij}} M_{pq}^{\textnormal{fine}} + (1 - \lambda) \cdot \frac{1}{|\mathcal{R}_{ij}|} \sum_{(p,q) \in \mathcal{R}_{ij}} M_{pq}^{\textnormal{fine}}
\end{equation}
Here, ${M}_{ij}^{\textnormal{coar}}$ represents the mixed pooling output value for the rectangular region $\mathcal{R}_{ij}$ within the fine map, which corresponds to the area uncertainty at the same position in the coarse map. $M_{pq}^{\textnormal{fine}}$ denotes the area uncertainty at position $(p,q)$ within $\mathcal{R}_{ij}$, and $|\mathcal{R}_{ij}|$ is the count of grids contained therein. The parameter $\lambda$ is a scaling factor that balances the contributions of maximum and average pooling.

The corresponding coarse maps for Fig. \ref{fig4}(a) and \ref{fig4}(b) are illustrated in \ref{fig4}(c) and \ref{fig4}(d), with a grid resolution of $80m$. Equipped with both the coarse and fine maps, we can now delve into the Coarse2fine A* algorithm for global path planning.

\subsection{Global Planning via Coarse2fine A* Algorithm}
The core of A* algorithm lies in the design of the cost function, which is expressed as:
\begin{equation}
\label{eqn6}
    F(n)=G(n)+H(n)
\end{equation}
where $F(n)$ represents the aggregate cost of node $n$; $G(n)$ and $H(n)$ are the current cost and the heuristic cost respectively. In the traditional A* algorithm, $G(n)$ is the measure of the actual path length from the start to node $n$. Expanding upon this, we incorporate an uncertainty factor as follows:
\begin{equation}
\label{eqn7}
    G (\overrightarrow{nn'})= d_{\overrightarrow{nn'}} \cdot \alpha_{u}(n')
\end{equation}

\begin{equation}
\label{eqn8}
    \alpha_{u} (n')=\frac{1}{1-U(n')}
\end{equation}
where $d_{\overrightarrow{nn'}}$  is the distance between $n$ and $n'$; $U(n')$ denotes the area uncertainty associated with node $n'$. As $U(n')$ increases, the current cost $G (\overrightarrow{nn'})$ correspondingly increases, encouraging the UGV to navigate through nodes with lower risk. The A* algorithm's criterion for optimality requires that $H(n)$ should not exceed the true cost from node n to the destination. Considering that $\alpha_u (n')$ surpasses 1, and $H(n)$ is predicated on Euclidean distance, the prerequisites for maintaining optimality remain upheld.

Beyond refining the cost function, the Coarse2fine A* primarily focuses on boosting the algorithm's efficiency. The basic idea is hierarchical planning, employing uncertainty maps with different resolutions at different layers, facilitating a transition from broad-scale planning at higher layer to detailed planning at finer layer.  This approach guarantees that, within each layer, the A* algorithm adjusts its search granularity according to the map's scale—utilizing a stride of $80m$ on a $2000m\times2000m$ map, $10m$ on an $80m\times80m$ map, and $1m$ on a $10m\times10m$ map. Concurrently, among different grids within each layer, the A* algorithm can run in parallel using multiple threads.

The integration of adaptive search granularity and parallel processing forms the core of the Coarse2fine A* algorithm, significantly improving its computational efficiency while maintaining the integrity of the pathfinding outcomes.

\begin{figure}
    \centering
    \includegraphics[width=0.9\linewidth]{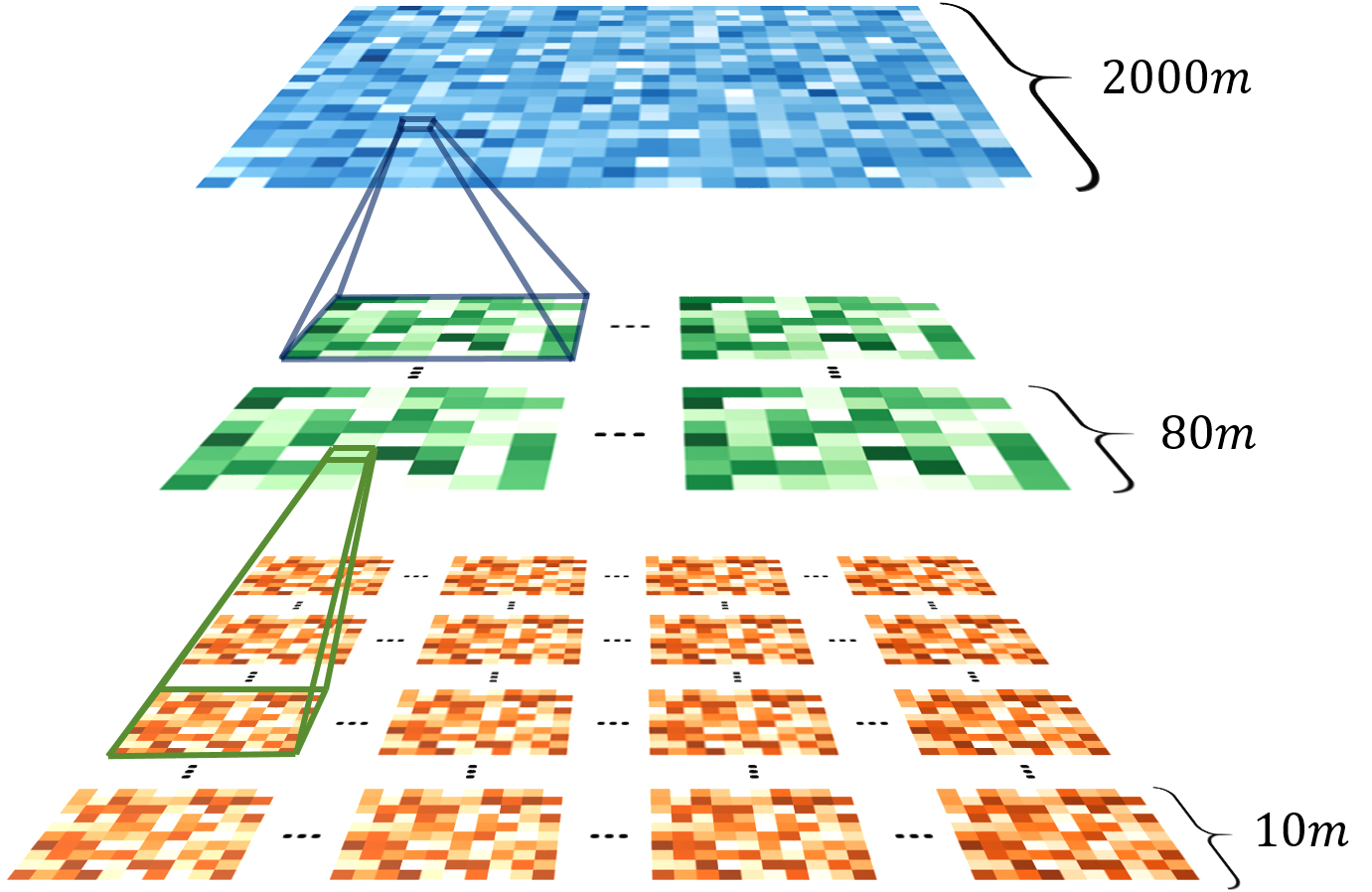}
    \caption{The visualization of the Coarse2fine A* algorithm.}
    \label{fig5}
\end{figure}

The architecture permits the flexible determination of the number of layers and the resolution specific to each layer, tailored to meet distinct operational demands. In this paper, a three-layered Coarse2fine framework is implemented on a $2000m\times2000m$ map, with each layer characterized by resolutions of $80m$, $10m$, and $1m$, respectively. The visualization of the Coarse2fine A-star algorithm is illustrated in Fig. \ref{fig5}, and the entire process is presented in Algorithm \ref{algorithm1}.

\begin{algorithm}[htbp]
    \small
    \label{algorithm1}
    \caption{Coarse2fine A* Algorithm}
    \SetKwInOut{Input}{Input}
    \SetKwInOut{Output}{Output}
    \SetKwFunction{F}{F}
    \SetKwFunction{COST}{COST}
    \SetKwFunction{GETNEIGHBOR}{GETNEIGHBOR}
    \SetKwFunction{H}{H}
    \SetKwFunction{EXTRACTPATH}{EXTRACTPATH}
    \SetKwFunction{FMain}{A*Search}
    \SetKwProg{Fn}{function}{:}{}

    \Input{start node $n_s$, goal node $n_g$, fine map $M^{\textnormal{fine}}$}
    \Output{list ${\textnormal{Path}}^{\textnormal{fine}}$}
    \BlankLine
    Initialize: stride $s^{\textnormal{fine}}$, $s^{\textnormal{midd}}$, $s^{\textnormal{coar}}$, middle map $M^{\textnormal{midd}}$ and coarse map $M^{\textnormal{coar}}$ with mixed pooling\;
    run \FMain{{$n_s$, $n_g$, $s^{\textnormal{coar}}$, $M^{\textnormal{coar}}$}} $\rightarrow$ list ${\textnormal{Path}}^{\textnormal{coar}}$\;
    \For{\textnormal{node} $n_{i}^{\textnormal{coar}}$ \textnormal{in} ${\textnormal{Path}}^{\textnormal{coar}}$}
    {
        \FMain{{$n_{i}^{\textnormal{coar}}$, $n_{i,\textnormal{next}}^{\textnormal{coar}}$, $s^{\textnormal{midd}}$, $M^{\textnormal{midd}}$}} on a thread \\ $\rightarrow$ list ${\textnormal{Path}}_{i}^{\textnormal{midd}}$\;
        merge all the ${\textnormal{Path}}_{i}^{\textnormal{midd}}$ $\rightarrow$ list ${\textnormal{Path}}^{\textnormal{midd}}$\;
    }
    \For{\textnormal{node} $n_{i}^{\textnormal{midd}}$ \textnormal{in} ${\textnormal{Path}}^{\textnormal{midd}}$}
    {
        \FMain{{$n_{i}^{\textnormal{midd}}$, $n_{i,\textnormal{next}}^{\textnormal{midd}}$, $s^{\textnormal{midd}}$, $M^{\textnormal{midd}}$}} on a thread \\ $\rightarrow$ list ${\textnormal{Path}}_{i}^{\textnormal{fine}}$\;
        merge all the ${\textnormal{Path}}_{i}^{\textnormal{fine}}$ $\rightarrow$ list ${\textnormal{Path}}^{\textnormal{fine}}$\;
    }
    \KwRet ${\textnormal{Path}}^{\textnormal{fine}}$
    \BlankLine
    \Fn{\FMain{$n_1$, $n_2$, $ns$, $M$}}{
        Initialize: priority queue $\textnormal{OpenList}$, list $\textnormal{ClosedList}$, dictionary $\textnormal{Parent}$, dictionary $G$\;
        $G(n_1) = 0; G(n_2) = \infty$\;
        push $(F(n_1), n_1)$ into $\textnormal{OpenList}$\;
        \While{\textnormal{$\textnormal{OpenList}$ is not empty}}
        {
            pop node $n$ from $\textnormal{OpenList}$ into $\textnormal{ClosedList}$\; 
            \eIf{$n == n_2$} {break\;}
            {
                \For{\textnormal{neighbor node} $n'$ \textnormal{in} \GETNEIGHBOR$(n,s)$}
                {
                    new cost $= G(n)$ + \COST$(n, n')$\;
                    \If{$n'$ is not in $G$}{$G(n')=\infty$\;}
                    \If{\textnormal{new cost} $<G(n')$} {$G(n')=$ new cost and $\textnormal{Parent}(n')=n$ and push $(F(n'),n')$ into $\textnormal{OpenList}$\;}
                }
            }
        }
        \EXTRACTPATH$(\textnormal{Parent}) \rightarrow \textnormal{PathList}$\;
        \KwRet $\textnormal{PathList}$
    }
\end{algorithm}
\subsection{Path Smoothing via Quadratic Programming}
We persist in utilizing the path smoothing approach proposed in \cite{jiang2023global}, modeling it as a QP problem and subsequently solving it. Three aspects are considered in the objective function of the QP problem: the smoothness, uniformity and compactness, and geometric similarity. These elements are quantified through respective cost components, namely, $\textnormal{cost}_1$, $\textnormal{cost}_2$, $\textnormal{cost}_3$, which are calculated by:
\begin{equation}
\label{eqn9}
    \textnormal{cost}_1 = \sum_{i=2}^{n-1} \Bigl((x_{i-1}+x_{i+1}-2x_i)^2 + (y_{i-1}+y_{i+1}-2y_i)^2\Bigr)
\end{equation}
\begin{equation}
\label{eqn10}
    \textnormal{cost}_2 = \sum_{i=1}^{n-1} \Bigl((x_i-x_{i+1})^2 + (y_i-y_{i+1})^2\Bigr)
\end{equation}
\begin{equation}
\label{eqn11}
    \textnormal{cost}_3 = \sum_{i=1}^{n} (x^2_i+y^2_i) + \sum_{i=1}^{n} (-2x_{ir}x_i-2y_{ir}y_i)
\end{equation}
In the computation of $\textnormal{cost}_3$, the constant term is disregarded. The optimization variables are the coordinates of the path nodes, denoted as $n_i (x_i,y_i)$, with $n_{ir} (x_{ir},y_{ir})$ representing the path node coordinates obtained through the Coarse2fine A* algorithm. These latter coordinates serve as a reference for the path smoothing module. The cumulative cost is formulated as a weighted sum of the three distinct costs:
\begin{equation}
\label{eqn12}
    \textnormal{cost} = \omega_1 \textnormal{cost}_1+\omega_2 \textnormal{cost}_2+\omega_3 \textnormal{cost}_3
\end{equation}

The cost function can be subtly modified to conform to the standard form of the QP problem. Denote the optimization variables as:
\begin{equation*}
    X=[x_1,y_1,x_2,y_2,...,x_n,y_n ]^T
\end{equation*}
Subsequently, the cost function can be reformulated as:
\begin{equation}
\label{eqn13}
    cost =X^T(\omega_1 A^T_1 A_1 + \omega_2 A^T_2 A_2 + \omega_3 I)X + \omega_3 f^T X
\end{equation}
where
\begin{gather*}
    A^T_1 = 
    \begin{bmatrix}
    1 & & & & \\
    0 &1 & & & \\
    -2 &0 &\ddots & & \\
    0 &-2 &\ddots & & \\
    1 &0 &\ddots & & \\
     &1 &\ddots & & \\ 
     & &\ddots & & \\\notag
    \end{bmatrix} 
     ,
    A^T_2 = 
    \begin{bmatrix}
    1 & & & &  \\
    0 &1 & & & \\
    -1 &0 &\ddots & & \\
    0 &-1 &\ddots & & \\
    1 &0 &\ddots & & \\
     &1 &\ddots & & \\ 
     & &\ddots & & \\\notag
    \end{bmatrix} 
     , \notag \\
     f=[-2x_{1r},-2y_{1r},-2x_{2r},-2y_{2r},...,-2x_{nr},-2y_{nr}]^T. \notag 
\end{gather*}
Here, the dimensions of $A^T_1$ and $A^T_2$ are (2n×2n-4) and (2n×2n-2) respectively.

Concerning the constraints pertinent to the path smoothing problem, we primarily consider that the coordinates of each path node are restricted to deviate within a specific boundary around the reference coordinates to guarantee safety. Assuming the allowable deviation in each direction is represented as $s^{fine}$, and the reference coordinates are denoted by $X_r=[x_{1r},y_{1r},x_{2r},y_{2r},...,x_{nr},y_{nr}]^T$, the final formulation of the QP problem is given by:
\begin{equation}
\label{eqn14}
    \min \quad X^T(\omega_1 A^T_1 A_1+\omega_2 A^T_2 A_2+\omega_3 I)X+\omega_3 f^T X
\end{equation}

\begin{equation}
\begin{aligned}
    s.t. \quad &X_r- \frac{s_s}{2}[1,1,...,1]^T \leq X \leq X_r + \frac{s_s}{2}[1,1,...,1]^T \\
    & n_1=n_{1r}\\
    & n_n=n_{nr}\notag
\end{aligned}
\end{equation}
which can be effectively solved using various readily available solvers, such as qpsolvers.

\begin{figure}[htbp]
  \centering
  \begin{minipage}{0.48\linewidth}
    \centering
    \includegraphics[width=1\linewidth]{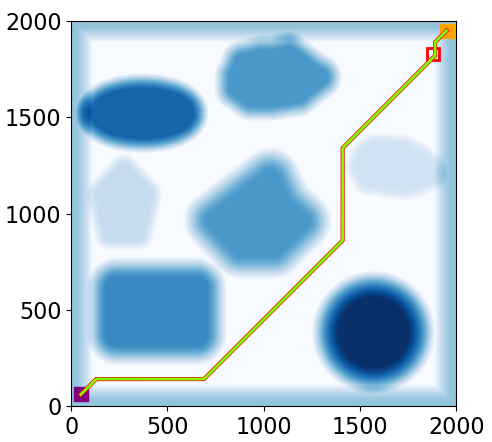} 
    \centerline{(a)}
  \end{minipage}
  \begin{minipage}{0.48\linewidth}
    \centering
    \includegraphics[width=1\linewidth]{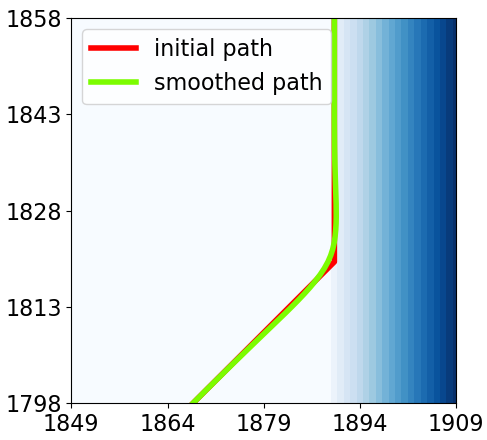} 
    \centerline{(b)}
  \end{minipage}
  \caption{The global path obtained by Coarse2fine A* algorithm (a) and its smoothing application (b).}
  \label{fig6}
\end{figure}

The path generated via the coarse2fine A* algorithm typically consists of over 1000 nodes. Directly feeding such an extensive path into the path smoothing module can lead to considerable computational demand and prolonged processing time. To mitigate this, we implement a rolling optimization strategy in conjunction with a path stitching approach, where only $n_o$ nodes are input into the QP solver each time. During each iteration, the end node from the preceding iteration is chosen as the reference node. Additionally, $n_b$ nodes in the backward direction (which have already been optimized), and $n_f=n_o-n_b-1$ nodes in the forward direction (awaiting optimization) are incorporated as the solver's input. To seamlessly integrate the path segments, it is essential to keep certain nodes at the forefront with fixed coordinates, while the terminal node's coordinates are maintained constant to guarantee arrival at the destination. The coordinates of other nodes are adaptable, subject to the optimization objectives. In cases where the residual number of unoptimized nodes falls short of $n_f$, the destination is assigned as the end node, complemented with $n_o-1$ nodes forward, and the QP problem is resolved using the same methodology. Algorithm \ref{algorithm2} highlights the primary steps of the path smoothing process. Fig. \ref{fig6}, within the designed scenario, exhibits the global path and its smoothing application. A comprehensive examination of the algorithm is set forth in Section \ref{Sec:Simu&Expe}.

\begin{algorithm}[htbp]
    \caption{Path Smoothing Algorithm via QP}
    \label{algorithm2}
    \KwIn{list ${\textnormal{Path}}^{\textnormal{fine}}$}
    \KwOut{list $\textnormal{SmoothedPath}$.}
    \BlankLine
    Initialize: list $\textnormal{SmoothedPath}$\;
    \For{\textnormal{node list} $N_{\textnormal{list}}$ \textnormal{with} $n_o$ \textnormal{nodes in} ${\textnormal{Path}}^{\textnormal{fine}}$}
    {
        determine whether the first or the last or other iteration\; 
        model the QP problem\; 
        solve the QP problem $\rightarrow$ list $\textnormal{SmoothedNodeList}$\; 
        append $\textnormal{SmoothedNodeList}$ into $\textnormal{SmoothedPath}$\; 
    }
    \KwRet $\textnormal{SmoothedPath}$
\end{algorithm}
\section{Local Planning} \label{Sec:Local Planning}
\begin{figure}
    \centering
    \includegraphics[width=0.7\linewidth]{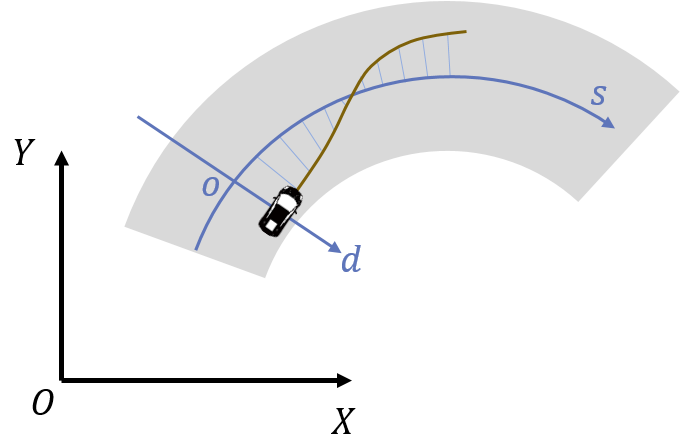}
    \caption{Frenet coordinate system.}
    \label{fig7}
\end{figure}

After obtaining the path as introduced in section \ref{Sec:GlobalPlanning}, it is employed as a reference path for local planning with method of deterministic sampling. The local planning process is primarily executed within the Frenet coordinate system, as illustrated in Fig. \ref{fig7}, wherein the reference path constitutes the foundation for establishing the coordinate system.

\subsection{Candidate Trajectories Generation}
In the Frenet coordinate system sod, it becomes feasible to decouple the lateral (perpendicular to the reference path) and longitudinal (along the reference path) motions. Planning is performed independently for each, represented by $d(t)$ for lateral and $s(t)$ for longitudinal movements. Their combination yields a complete trajectory.

As mentioned previously, local trajectory planning is essentially addressing an Optimal Control Problem (OCP). The deterministic sampling method seeks to approximate an optimal solution within the OCP's feasible set through a process of sampling, constraints checking, and cost evaluation.

For trajectory planning of UGV, the sampling space is $[t, d, \dot{d}, \ddot{d}, s, \dot{s}, \ddot{s}]$, denoted as $\mathcal{S}$. However, not all variables in $\mathcal{S}$ need to be sampled. In each iteration, a reference target state $\mathcal{S}_r = [t_r, d_r, \dot{d}_r, \ddot{d}_r, s_r, \dot{s}_r, \ddot{s}_r]$ is determined first. Around $\mathcal{S}_r$, the variables that need to be sampled vary within a certain range to derive all possible sampled states. Practically, sampling is executed within a subspace, $[t, d, s]$ of $\mathcal{S}$, abbreviated as $\mathcal{S}'$, indicating that merely these three variables are subject to variation, while the rest are held constant. Given the UGV's predetermined target speed $v_r$, the target state $\mathcal{S}_r$ is specified as follows: $t_r$ remains an adjustable parameter for determination, $s_r=v_c t_r$ and $v_c$ is the current speed, $\dot{s}_r=v_r$, $\ddot{s}_r$, $d_r$, $\dot{d}_r$ and $\ddot{d}_r$ are set to 0. This signifies the intent for the UGV to advance along the reference path by a distance of $v_c t_r$ post $t_r$, aiming to achieve a relatively stable state devoid of lateral motion and longitudinal acceleration. It is noteworthy that hitting the target state $\mathcal{S}_r$ is not obligatory; instead, the state with the lowest trajectory cost from the generated candidates within its sampling range is selected. Therefore, the presence of obstacles near the target state does not inherently lead to potential dangers.

\begin{figure}
    \centering
    \includegraphics[width=1.0\linewidth]{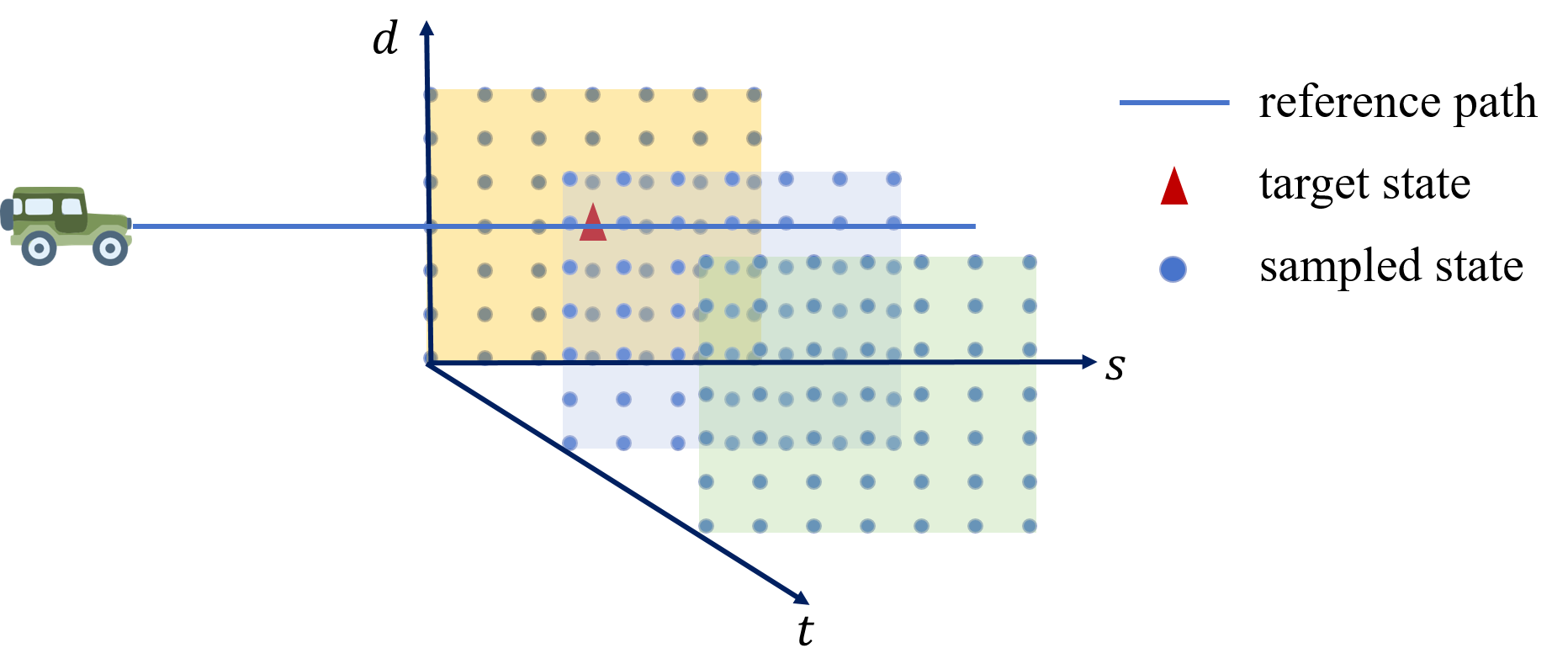}
    \caption{Example of sampling in $\mathcal{S}'$.}
    \label{fig8}
\end{figure}

In the practical sampling procedure, we initially define the sampling range from $\mathcal{S}'_{\min} = [t_{\min},d_{\min},s_{\min}]$ to $\mathcal{S}'_{\max} = [t_{\max},d_{\max},s_{\max}]$. Subsequently, it is divided into two segments $[\mathcal{S}'_{\min},\mathcal{S}'_r)$ and $(\mathcal{S}'_r,\mathcal{S}'_{\max}]$, where uniform sampling is conducted within each segment. The predetermined number of samples in these two intervals is represented by $N_l$ and $N_u$, respectively. Taking $d$ as an example, sampling is performed in the range $[d_{\min},d_r)$:
\begin{equation}
\label{eqn15}
    \{d_i | d_i = d_r \frac{N_l - i}{N_l} + d_{\min} \frac{i}{N_l}, \quad i = 1, 2, \ldots, N_l\}
\end{equation}
A similar method is adopted for sampling in the interval $(d_r,d_{\max}]$. The target state $\mathcal{S}_r$ is always sampled, thus the total number of samples for each variable in $\mathcal{S}'$ is $N_l+N_u+1$. Fig. \ref{fig8} illustrates the sampling outcomes at a specific moment in local planning.

The sampled trajectory in Frenet coordinate system emerges from the integration of $d(t)$ and $s(t)$. Typically, the control module receives discrete trajectory points as input. Hence, $d(t)$ and $s(t)$ are discretized at intervals of $\Delta t$. Subsequently, $d$ and $s$ can be merged with the reference path point by point, transitioning the trajectory from the Frenet to the global coordinate system.

\begin{figure}
    \centering
    \includegraphics[width=0.8\linewidth]{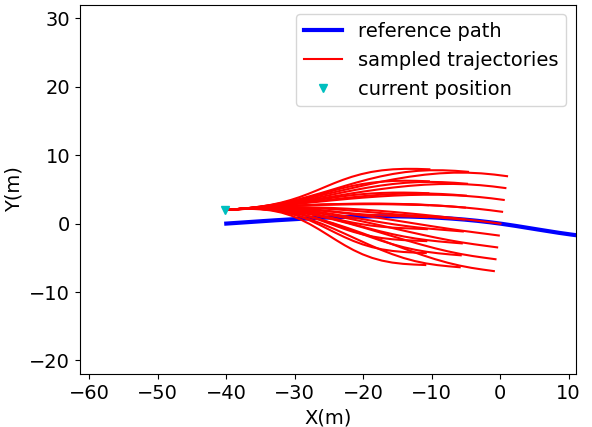}
    \caption{The sampled trajectories.}
    \label{fig9}
\end{figure}

In practical applications, the number of samples influences both the trajectory's quality and the planning efficiency. Generally, a larger sample count yields a trajectory of superior quality approaching optimality but also increases computational complexity and, consequently, processing time. The selection of the sampling number should be aligned with the computational capabilities of the UGV's control device. Drawing from our simulations and real-world deployment examinations, we provide two sampling number alternatives, listed in Table \ref{table1} along with other parameters. The sampled trajectories are shown in Fig. \ref{fig9} (out of a total of 315 trajectories obtained with sampling number option 1, only 27 are displayed for the sake of clarity).

\begin{table}[ht]
\centering
\caption{Parameters of Candidate Trajectories Generation}
\label{table1}
\begin{tabular}{@{}ccccccc@{}}
\toprule
Variable & \shortstack{Sampled \\ or Not} & $\mathcal{S}_{\min}$ & $\mathcal{S}_{\max}$ & $N_l$ & $N_u$ & \shortstack{Default \\ Value} \\ 
\midrule
$t_t$ & $\surd$ & 3s & 7s & 2/3 & 2/4 & $t_r=5s$ \\
$d_t$ & $\surd$ & -7m & 7m & 4/7 & 4/7 & $d_r=0$ \\
$\dot{d}_t$ & $\times$ & - & - & - & - & $\dot{d}_r=0$ \\
$\ddot{d}_t$ & $\times$ & - & - & - & - & $\ddot{d}_r=0$ \\
$s_t$ & $\surd$ & 0.8$s_r$ & 1.2$s_r$ & 3/4 & 3/4 & $s_r=t_r v_c$ \\
$\dot{s}_t$ & $\times$ & - & - & - & - & $\dot{s}_r=v_r$ \\
$\ddot{s}_t$ & $\times$ & - & - & - & - & $\ddot{s}_r=0$ \\
$v_r$ & - & - & - & - & - & 25km/h \\
$\Delta t$ & - & - & - & - & - & 0.25s \\
\midrule
\multicolumn{7}{l}{Total number option 1: $5\times9\times7=315$} \\
\multicolumn{7}{l}{Total number option 2: $7\times15\times9=945$} \\
\bottomrule
\end{tabular}
\end{table}

\subsection{Constrains Checking}
Not all trajectories will be considered in the cost evaluation module; prior to that, they are subjected to constraints checking to exclude those that fail to satisfy the UGV's dynamic and safety constraints, which mainly include the following five parts:
\begin{enumerate}
    \item UGV’s velocity $v$ within $[v_{\min}, v_{\max}]$,
    \item longitudinal acceleration $a_y$ within $[a_{y,\min}, a_{y,\max}]$,
    \item lateral acceleration $a_x$ within $[a_{x,\min}, a_{x,\max}]$,
    \item curvature of trajectory $c$ within $[c_{\min}, c_{\max}]$,
    \item Safety indicator $e$ within $[0, e_{\textnormal{thld}}]$,
\end{enumerate}
A point-wise inspection method is employed to evaluate each trajectory's compliance with these constraints. If any point fails to satisfy any constraint, the entire trajectory is considered non-compliant and thus discarded.

Since the trajectory is planned within the Frenet coordinate system, additional computations are required to derive the numerical values necessary for checking constraints 1) to 4). Velocity $v$, longitudinal and lateral acceleration $a_y$ and $a_x$, curvature $c$ of each point on the planned trajectory can be calculated by:
\begin{align}
    c_i &= \frac{\theta_{i+1} - \theta_i}{s_{i+1} - s_i} \\
    v_i &= \sqrt{\dot{s_i}^2 (1 - c_i d_i)^2 + \dot{d_i}^2} \\
    a_{y,i} &= \frac{v_{i+1} - v_i}{\Delta t} \\
    a_{x,i} &= v_i^2 c_i
\end{align}
where $i$ indicates the $i$-th point on the trajectory. For $a_{y,i}$, it can also be approximated by $\ddot{s_i}$ to reduce computation load.

\begin{figure}
    \centering
    \includegraphics[width=0.8\linewidth]{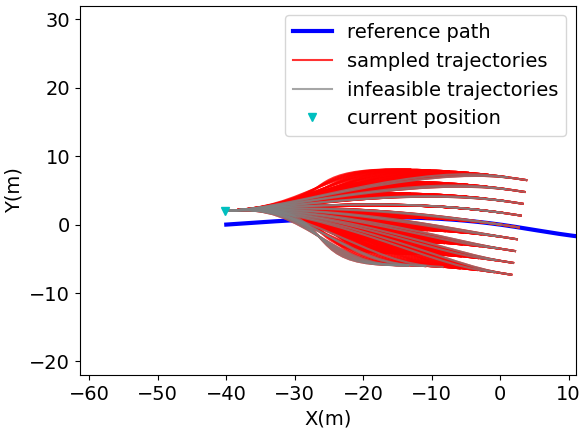}
    \caption{The infeasible trajectories among candidates.}
    \label{fig10}
\end{figure}

For constraint 5), the trajectory is inputted into the potential fields generated by all obstacles, and the overall field value at each point along the trajectory is computed. If the value at any point surpasses the threshold $e_{\textnormal{thld}}$, the trajectory is deemed infeasible. Compared to traditional collision avoidance constraints, employing the safety margin method by potential field allows for an earlier awareness of obstacles' influence, thereby enhancing the safety of the planned trajectory. In the scenario depicted in Fig. \ref{fig9}, 26 trajectories are discarded during constraints checking, as illustrated in Fig. \ref{fig10}. The parameters employed in constraints checking are presented in Table \ref{table2}.

\begin{table}[ht]
\centering
\caption{Parameters of Constrains Checking}
\label{table2}
\begin{tabular}{cc}
\toprule
Variable & Value \\
\midrule
$v_{\min} / v_{\max}$ & 0 / 50 km/h \\
$a_{y,\min} / a_{y,\max}$ & -7 / 3.5 m/s$^2$ \\
$a_{x,\min} / a_{x,\max}$ & -4 / 4 m/s$^2$ \\
$c_{\min} / c_{\max}$ & -0.43 / 0.43 \\
$e_{\textnormal{thld}}$ & 10 \\
\bottomrule
\end{tabular}
\end{table}

\subsection{Cost Evaluation}
For each feasible trajectory, its associated cost is determined by considering four key dimensions:

1) the smoothness of the trajectory: we measure it through the maximum lateral acceleration and the lateral and longitudinal jerk (approximated by $\dddot{s}$ and $\dddot{d}$ in Frenet coordinate system), that is,
\begin{equation}
J_s = \omega_a \left( \max_{i=1,\ldots,n} a_x \right) + \sum_{i=1}^{n} \left( \omega_s \dddot{s}_i + \omega_d \dddot{s}_i \right)
\end{equation}
where $\omega_a$, $\omega_s$, $\omega_d$ are the weight coefficients.

2) the deviation between sampled terminal state $\mathcal{S}_t$ and reference target state $\mathcal{S}_r$. As $\mathcal{S}_t$ approaches $\mathcal{S}_r$, this cost term decreases, that is,
\begin{equation}
J_t = (\mathcal{S}_t - \mathcal{S}_r) \bm{\omega_t} (S_t - S_r)^T
\end{equation}
where $\bm{\omega_t}$ is a weight matrix with non-zero elements only along the diagonal.

3) the safety level of the trajectory, which is assessed by the safety indicator $e$ calculated within the constrains checking module:
\begin{equation}
J_e = \omega_e \cdot e
\end{equation}
where $\omega_e$ is the weight coefficient.

4) the consistency of the trajectory: To mitigate potential instability in the UGV’s movement caused by substantial differences between trajectories across successive iterations, this dimension ensures the planned trajectory remains within a permissible variance from the prior optimal trajectory. This is quantified by the divergence in terminal states, with the prior optimal trajectory's terminal state denoted as $\mathcal{S}_b$, and the related cost computed as:
\begin{equation}
J_c = (\mathcal{S}_t - \mathcal{S}_b) \bm{\omega_c} (\mathcal{S}_t - \mathcal{S}_b)^T
\end{equation}
where $\bm{\omega_c}$ is the weight matrix.

\begin{figure}
    \centering
    \includegraphics[width=0.8\linewidth]{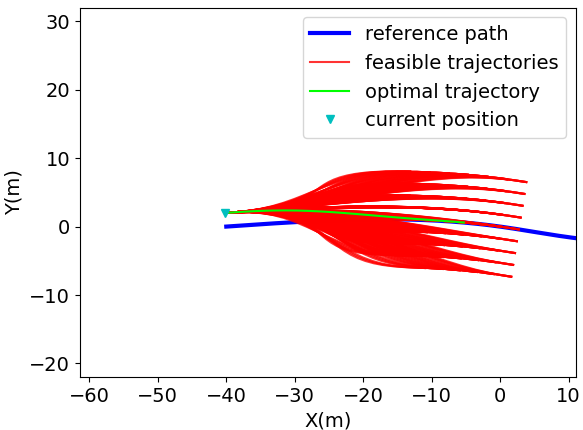}
    \caption{The infeasible trajectories among candidates.}
    \label{fig11}
\end{figure}

Combining the above costs yields the total cost for each trajectory:
\begin{equation}
J = J_s + J_t + J_e + J_c
\end{equation}
Subsequently, the trajectory with the minimum cost is selected from all feasible trajectories, as shown in Fig. \ref{fig11}. The parameters utilized in cost computation are listed in Table \ref{table3}. The entire process of the deterministic sampling algorithm is illustrated in Algorithm \ref{algorithm3}.

\begin{table}[ht]
\centering
\caption{Parameters of Cost Evaluation}
\label{table3}
\begin{tabular}{cc}
\toprule
Variable & Value \\
\midrule
$\omega_a$ & 1 \\
$\omega_s$ & 2 \\
$\omega_d$ & 5 \\
$\bm{\omega_t}$ & diag(5,20,0,0,18,0,0) \\
$\omega_e$ & 100 \\
$\bm{\omega_c}$ & diag(0,1.5,0,0,0.2,0,0) \\
\bottomrule
\end{tabular}
\end{table}

\begin{algorithm}[htbp]
    \caption{Deterministic Sampling Algorithm for Off-road Environment}
    \label{algorithm3}
    \KwIn{list $\textnormal{SmoothedPath}$, $\textnormal{GoalState}$}
    \BlankLine
    Initialize: list $\textnormal{UGVState}, \textnormal{ObstacleState}$\;
    \While{$\textnormal{UGVState}$ $\neq$ $\textnormal{ObstacleState}$}
    {
        update $\textnormal{ObstacleState}$\; 
        generate candidate trajectories in Frenet coordinate system $\rightarrow$ list $\textnormal{CandiTrajs}$\; 
        calculate $\textnormal{CandiTrajs}'$ information in global coordinate system\;
        check constrains of candidate trajectories to obtain feasible trajectories $\rightarrow$ list $\textnormal{FeasiTrajs}$\; 
        calculate the cost of all feasible trajectories\;
        obtain the trajectory with minimum cost $\rightarrow$ list $\textnormal{BestTraj}$\;
        upadte $\textnormal{UGVState}$\; 
    }
\end{algorithm}

\section{Simulation and Experiments} \label{Sec:Simu&Expe}
This section demonstrates the performance of the proposed algorithm in off-road scenarios. First, the quality and efficiency of the global planning algorithm are validated through simulations under both designed scenarios and randomly generated uncertainty environments. Subsequently, we assess the obstacle avoidance performance of the local planning algorithm in off-road environments. Finally, the algorithm's practical applicability is confirmed through its implementation on a real UGV platform.

\subsection{Validation of the Global Planning Module}
The implementation of the global planning algorithm was carried out on a laptop equipped with an Intel Core i7-13700H CPU, NVIDIA GeForce RTX 4060 GPU, and 32GB of RAM, operating under a Linux system.

\begin{figure}[htbp]
  \centering
  \begin{minipage}{0.48\linewidth}
    \centering
    \includegraphics[width=0.95\linewidth]{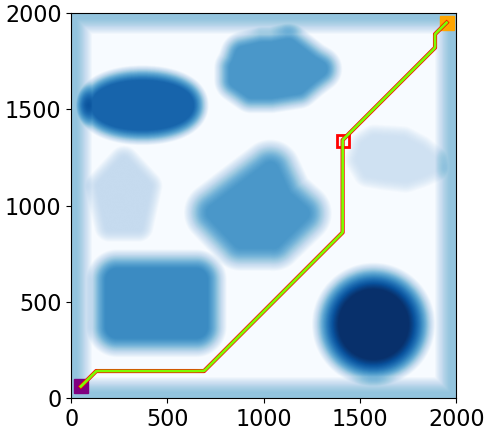}
    \vspace{5pt}
  \end{minipage}
  \begin{minipage}{0.48\linewidth}
    \centering
    \includegraphics[width=0.95\linewidth]{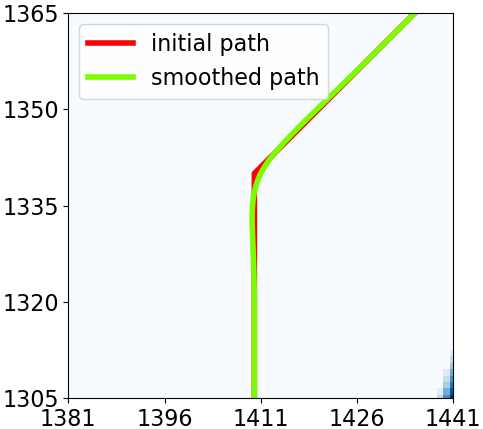}
    \vspace{5pt}
  \end{minipage}
  \vspace{5pt} 
  \centerline{(a) start node (50,60), end node (1950,1950)}
  \vspace{10pt} 
  \begin{minipage}{0.48\linewidth}
    \centering
    \includegraphics[width=0.95\linewidth]{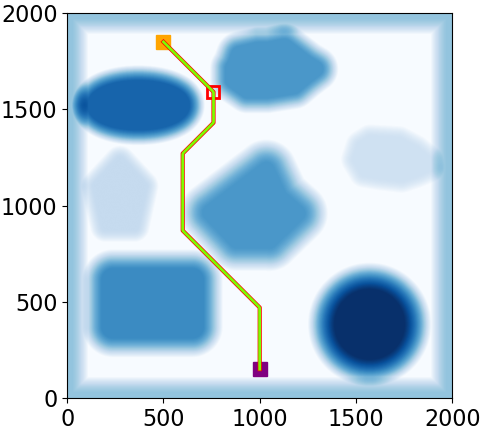}
  \end{minipage}
  \begin{minipage}{0.48\linewidth}
    \centering
    \includegraphics[width=0.95\linewidth]{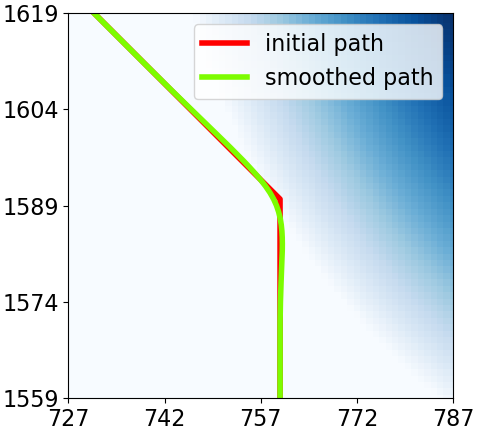} 
  \end{minipage}
  \vspace{5pt} 
  \centerline{(b) start node (1000,150), end node (500,1850)}
  \vspace{5pt} 
  \begin{minipage}{0.48\linewidth}
    \centering
    \includegraphics[width=0.95\linewidth]{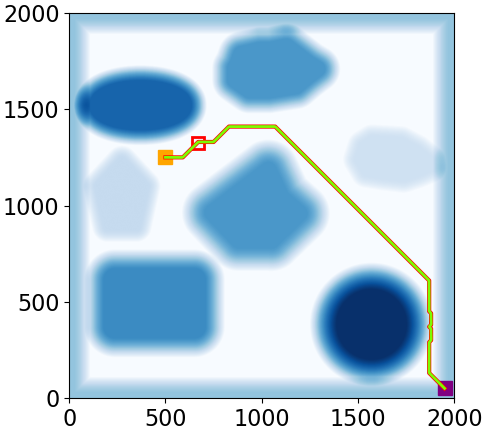} 
  \end{minipage}
  \begin{minipage}{0.48\linewidth}
    \centering
    \includegraphics[width=0.95\linewidth]{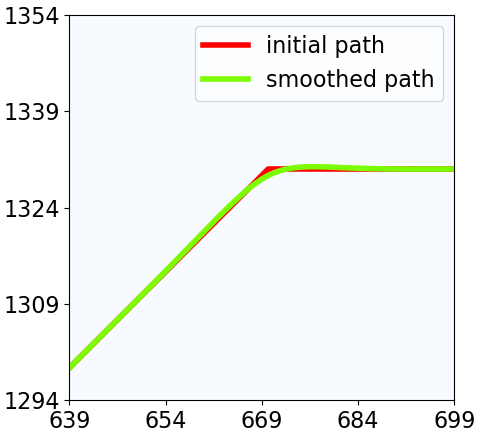} 
  \end{minipage}
  \centerline{(c) start node (1950,50), end node (500,1250)}
  \caption{The extracted path by Coarse2fine A* algorithm on designed map.}
  \label{fig12}
\end{figure}

We first execute the proposed Coarse2fine A* algorithm in the scenario depicted in Fig. \ref{fig2}, and the outcomes are shown in Fig. \ref{fig12}. The start node and the goal node are denoted by purple and orange squares respectively, while red hollow squares indicate local areas targeted for magnification to showcase the path smoothing effect. The algorithm is capable of rapidly responding to changes in the positions of the start and goal nodes.

\begin{figure}[htbp]
  \centering
  \begin{minipage}{0.48\linewidth}
    \centering
    \includegraphics[width=0.95\linewidth]{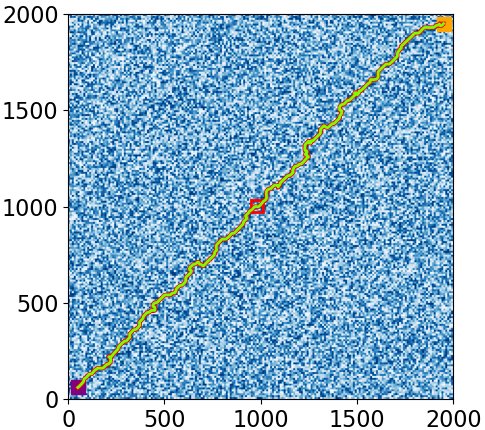}
    \vspace{5pt}
  \end{minipage}
  \begin{minipage}{0.48\linewidth}
    \centering
    \includegraphics[width=0.95\linewidth]{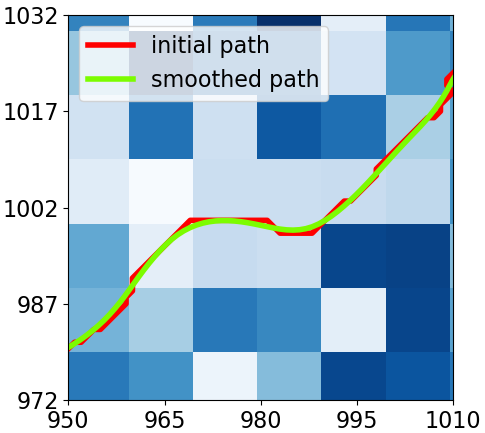}
    \vspace{5pt}
  \end{minipage}
  \vspace{5pt} 
  \centerline{(a) start node (50,60), end node (1950,1950)}
  \vspace{10pt} 
  \begin{minipage}{0.48\linewidth}
    \centering
    \includegraphics[width=0.95\linewidth]{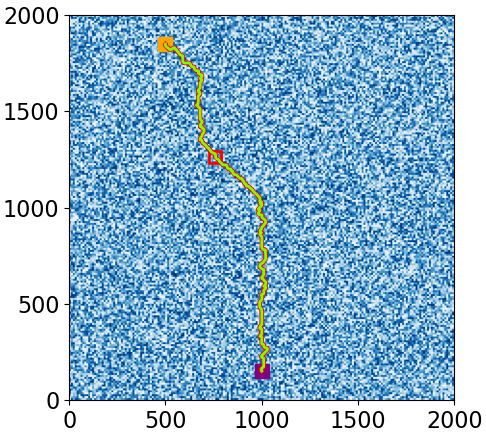}
  \end{minipage}
  \begin{minipage}{0.48\linewidth}
    \centering
    \includegraphics[width=0.95\linewidth]{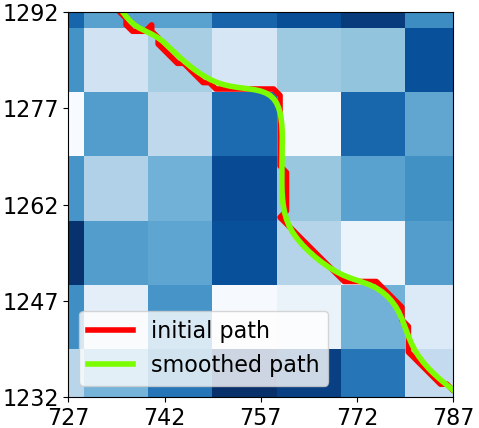} 
  \end{minipage}
  \vspace{5pt} 
  \centerline{(b) start node (1000,150), end node (500,1850)}
  \vspace{5pt} 
  \begin{minipage}{0.48\linewidth}
    \centering
    \includegraphics[width=0.95\linewidth]{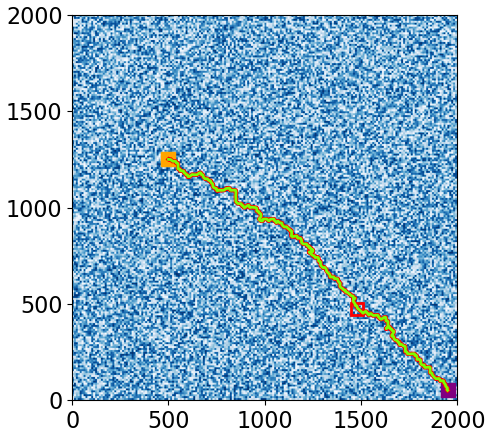} 
  \end{minipage}
  \begin{minipage}{0.48\linewidth}
    \centering
    \includegraphics[width=0.95\linewidth]{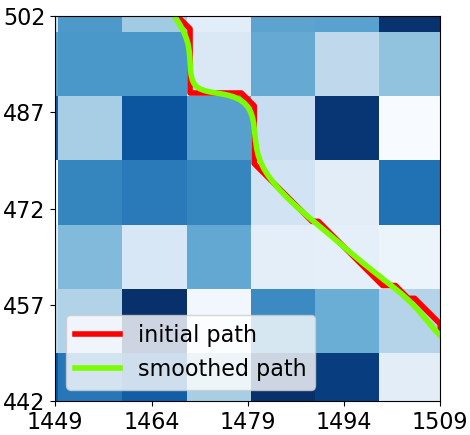} 
  \end{minipage}
  \centerline{(c) start node (1950,50), end node (500,1250)}
  \caption{The extracted path by Coarse2fine A* algorithm on designed map.}
  \label{fig13}
\end{figure}

Then we apply the algorithm on the randomly generated uncertainty map, facilitating a more effective comparative demonstration of the global planning module. The start node and goal nodes are also changed several times, with the planned paths being presented in Fig. \ref{fig13}.

\begin{figure}
    \centering \includegraphics[width=0.8\linewidth]{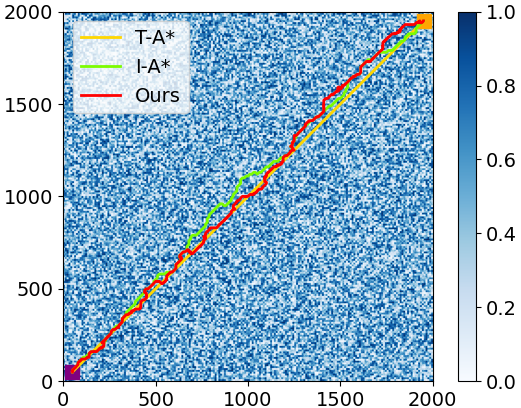}
    \caption{The paths obtained by different algorithms.}
    \label{fig14}
\end{figure}

From Fig. \ref{fig12}, it can be seen that the planned paths avoid high-risk areas and tend to pass through regions with lower uncertainty. This phenomenon becomes more evident in the zoomed-in view in Fig. \ref{fig13}, where, within the randomly generated uncertainty map, the planning results demonstrate the ability to navigate around high uncertainty areas in local regions. A numerical analysis further substantiates the risk-avoidance tendency of the coarse2fine A* algorithm. Across ten scenarios featuring variably positioned start and goal nodes on the randomly generated uncertainty map, planning is conducted using traditional A* (T-A*), the improved A* (I-A*) as mentioned in \cite{jiang2023global}, and Coarse2fine A* (Ours). The outcomes, as shown in Fig. \ref{fig14}, are summarized in Table \ref{table4}, with path uncertainty quantified as the mean uncertainty across all nodes constituting the path.

\begin{table}[htb]
\centering
\caption{Comparison of Planned Path Uncertainty with Different Algorithms}
\label{table4}
\begin{tabular}{@{}cccccc@{}}
\toprule
Start node & Goal node & T-A* & I-A* & Ours & Improvement \\
\midrule
(50,60) & (1950,1950) & 0.5060 & 0.2322 & 0.1875 & 19.25\% \\
(1000,150) & (500,1850) & 0.4793 & 0.3300 & 0.2077 & 37.06\% \\
(1950,50) & (500,1250) & 0.4757 & 0.4276 & 0.1874 & 56.17\% \\
(1750,1950) & (230,340) & 0.4844 & 0.2335 & 0.2015 & 13.70\% \\
(50,80) & (1850,20) & 0.5138 & 0.3060 & 0.2089 & 31.73\% \\
(210,560) & (1050,1940) & 0.4649 & 0.2490 & 0.1942 & 22.01\% \\
(1340,250) & (1940,1820) & 0.4700 & 0.2382 & 0.1773 & 25.57\% \\
(1030,930) & (1260,170) & 0.5221 & 0.3329 & 0.1800 & 45.93\% \\
(160,1340) & (1390,570) & 0.5046 & 0.3959 & 0.1921 & 51.48\% \\
(260,530) & (1730,840) & 0.5193 & 0.2617 & 0.1703 & 34.93\% \\
\midrule
\multicolumn{2}{c}{Average} & 0.4940 & 0.3007 & 0.1907 & 33.78\% \\
\bottomrule
\end{tabular}
\end{table}

Compared to baseline algorithms, the Coarse2fine A* algorithm achieves a notable average reduction in path uncertainty by over 30\%, significantly enhancing safety. This advancement is especially crucial in dangerous off-road scenarios.

Another critical aspect is the algorithm's efficiency.  In \cite{jiang2023global}, methods to boost efficiency during path smoothing process have been validated. This paper focuses exclusively on analyzing the time efficiency in the path planning stage. Experiments conducted under the same settings as in Table \ref{table4} reveal the time consumed by different algorithms (measured in seconds), as detailed in Table \ref{table5}. The results suggest that the Coarse2fine A* algorithm can further improve the efficiency of path planning slightly (as demonstrated in \cite{jiang2023global}, the efficiency of I-A* algorithm is already well-established), thus ensuring the algorithm's capability for real-time performance.
\begin{table}[ht]
\centering
\caption{Comparison of Time Consumed with Different Algorithms}
\label{table5}
\begin{tabular}{@{}cccccc@{}}
\toprule
Start node & Goal node & T-A* & I-A* & Ours & Improvement \\
\midrule
(50,60) & (1950,1950) & 0.5263 & 0.4346 & 0.3178 & 22.19\% \\
(1000,150) & (500,1850) & 10.6872 & 0.2945 & 0.2716 & 7.78\% \\
(1950,50) & (500,1250) & 7.4594 & 0.2095 & 0.2713 & -29.50\% \\
(1750,1950) & (230,340) & 3.7474 & 0.3473 & 0.2407 & 30.69\% \\
(50,80) & (1850,20) & 2.2051 & 0.1590 & 0.2467 & -55.16\% \\
(210,560) & (1050,1940) & 9.0845 & 0.2324 & 0.2382 & -2.50\% \\
(1340,250) & (1940,1820) & 10.3896 & 0.2222 & 0.2038 & 8.28\% \\
(1030,930) & (1260,170) & 2.0626 & 0.0687 & 0.1234 & -79.62\% \\
(160,1340) & (1390,570) & 7.0757 & 0.1823 & 0.2089 & -14.60\% \\
(260,530) & (1730,840) & 6.2015 & 0.2066 & 0.1966 & 4.84\% \\
\midrule
\multicolumn{2}{c}{Average} & 5.9440 & 0.2357 & 0.2319 & 1.62\% \\
\bottomrule
\end{tabular}
\end{table}

\subsection{Simulation of the Local Planning Module}
\begin{figure}[htbp]
  \centering
  \begin{minipage}{0.48\linewidth}
    \centering
    \includegraphics[width=1\linewidth]{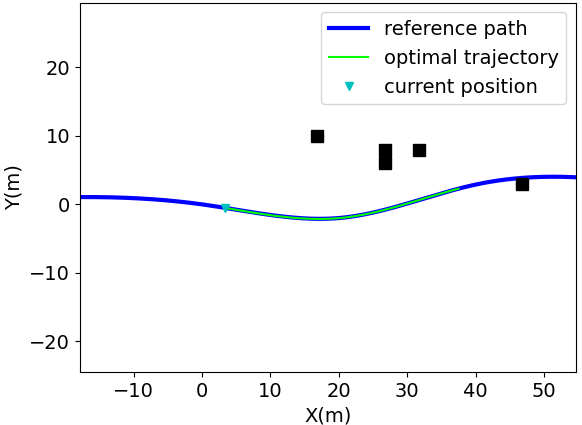} 
    \centerline{(a)}
  \end{minipage}
  \begin{minipage}{0.48\linewidth}
    \centering
    \includegraphics[width=1\linewidth]{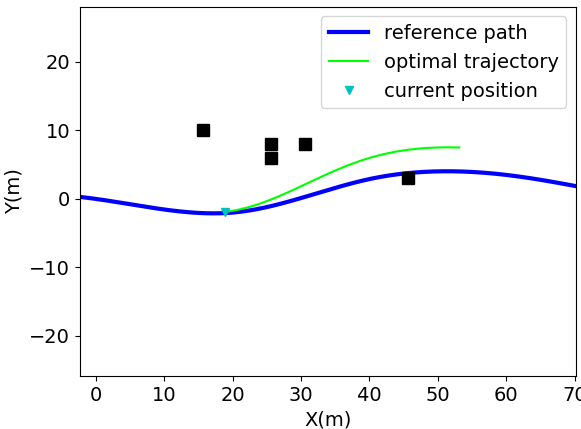} 
    \centerline{(b)}
  \end{minipage}
  \begin{minipage}{0.48\linewidth}
    \centering
    \includegraphics[width=1\linewidth]{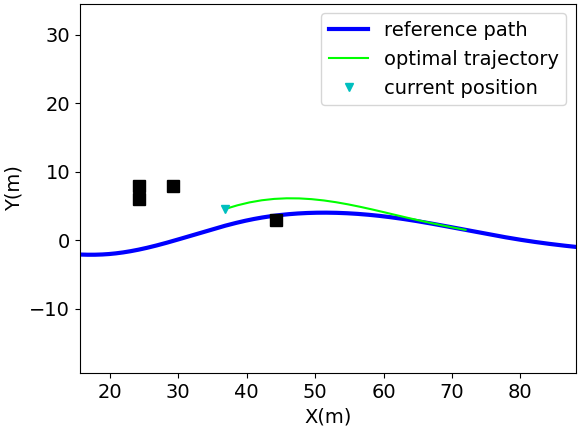} 
    \centerline{(c)}
  \end{minipage}
  \begin{minipage}{0.48\linewidth}
    \centering
    \includegraphics[width=1\linewidth]{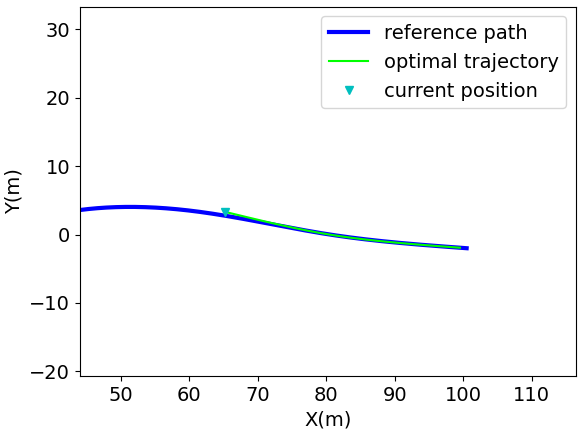} 
    \centerline{(d)}
  \end{minipage}
  \caption{The UGV avoids local obstacles during local planning.}
  \label{fig15}
\end{figure}

For the assessment of local planning, our primary focus lies on the performance of the UGV in avoiding local risks. To support the simulation experiments, we have supplemented a straightforward control module that updates the state of the UGV in the simulation environment after each planning iteration. Fig. \ref{fig15} illustrates the entire process through which the UGV successfully navigate along the reference path while simultaneously avoid local obstacles.

\begin{figure}[htbp]
  \centering
  \begin{minipage}{0.48\linewidth}
    \centering
    \includegraphics[width=1\linewidth]{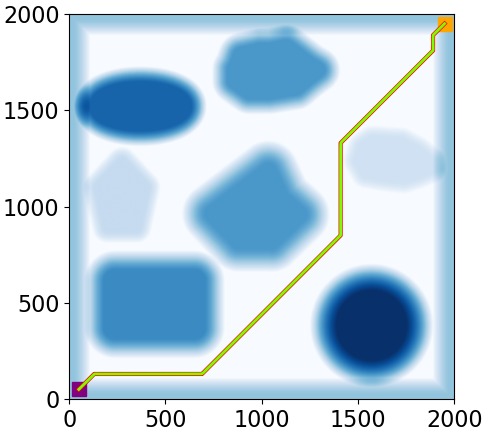}
  \end{minipage}
  \begin{minipage}{0.48\linewidth}
    \centering
    \includegraphics[width=0.90\linewidth]{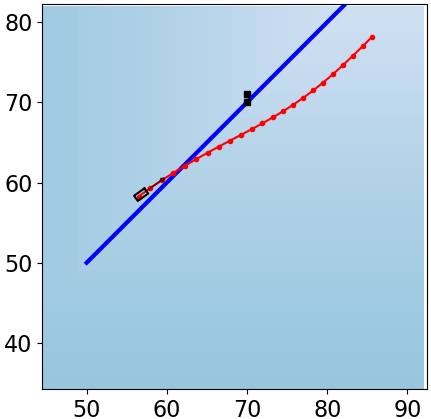}
  \end{minipage}
  \centerline{(a)}
  
  \begin{minipage}{0.48\linewidth}
    \centering
    \includegraphics[width=1\linewidth]{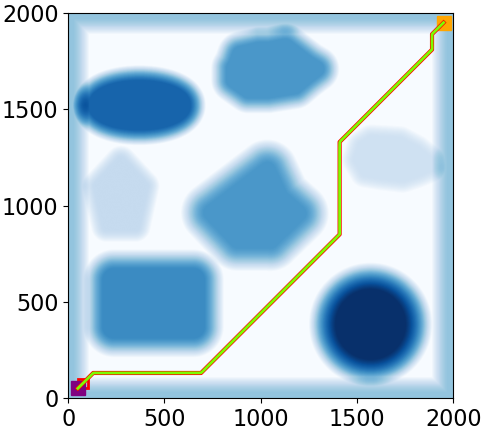}
  \end{minipage}
  \begin{minipage}{0.48\linewidth}
    \centering
    \includegraphics[width=0.91\linewidth]{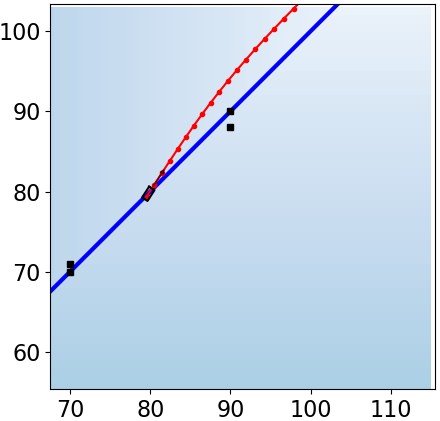} 
  \end{minipage}
  \centerline{(b)}
  \begin{minipage}{0.48\linewidth}
    \centering
    \includegraphics[width=1\linewidth]{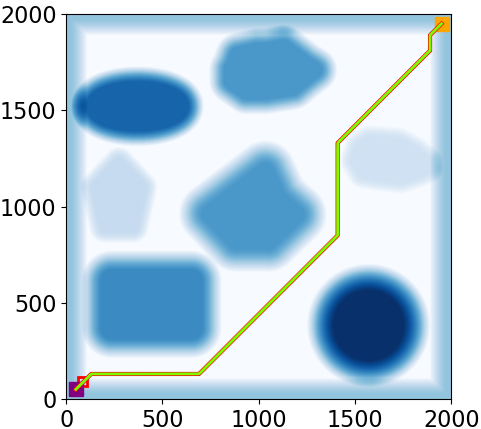} 
  \end{minipage}
  \begin{minipage}{0.48\linewidth}
    \centering
    \includegraphics[width=0.95\linewidth]{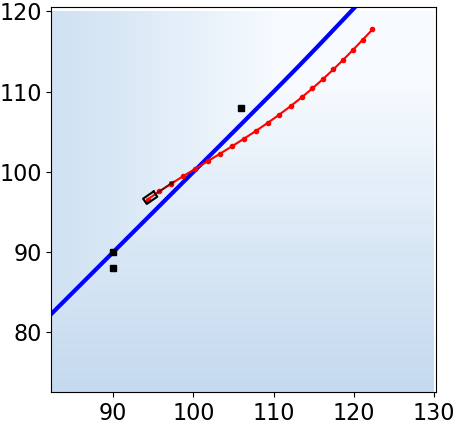} 
  \end{minipage}
  \centerline{(c)}
  \begin{minipage}{0.48\linewidth}
    \centering
    \includegraphics[width=1\linewidth]{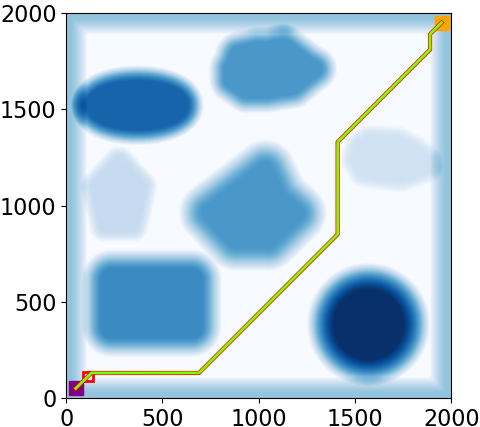} 
  \end{minipage}
  \begin{minipage}{0.48\linewidth}
    \centering
    \includegraphics[width=0.91\linewidth]{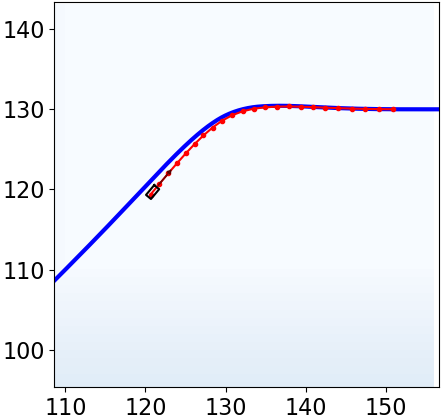} 
  \end{minipage}
  \centerline{(d)}
  \caption{The integrated simulation of global and local planning.}
  \label{fig16}
\end{figure}

The global and local planning modules have subsequently been integrated into a unified simulation environment, as demonstrated in Fig. \ref{fig16}. The simulation outcomes provide the basis for the real-world application of the algorithm on UGVs.

\subsection{Experimental Validation}
\begin{figure}
    \centering
    \includegraphics[width=1\linewidth]{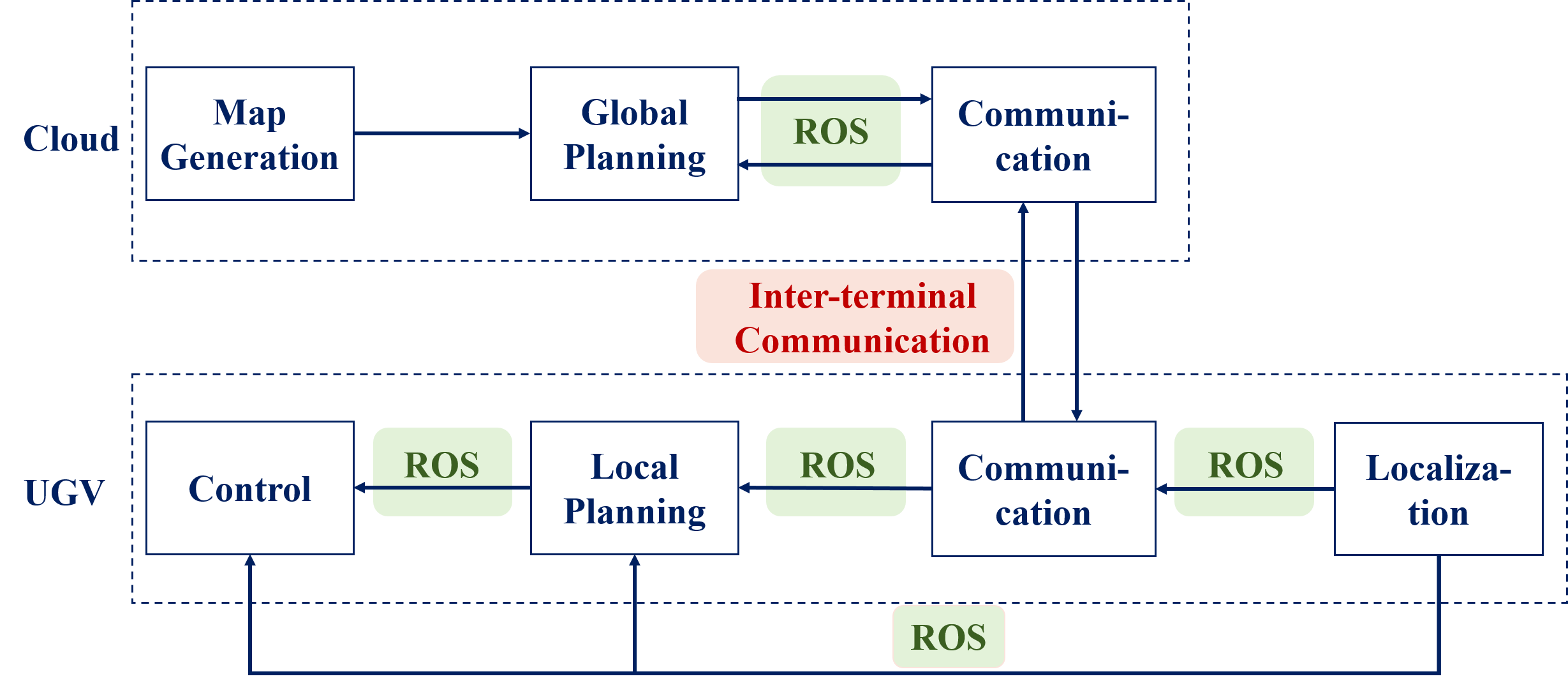}
    \caption{The communication architecture used in the real-world experiments.}
    \label{fig17}
\end{figure}

In the experimental validation, the local planning module was implemented on the UGV itself. As for the deployment of the global planning module, two alternatives were considered: the first is to deploy it also on the UGV, while the second is to deploy it on an additional control center, such as a cloud server. The former benefits from the absence of vehicle-to-cloud communication, whereas the latter is advantageous for centralized planning in multi-UGV scenarios. In practice, we opted the latter approach. Communication is a key element in real-world experimentation. If the algorithm is solely deployed on the UGV, internal communication amongst various system components is required, which can be efficiently managed through ROS (Robot Operating System). On the other hand, if adopting a cloud-and-UGV deployment scheme, inter-terminal communication becomes necessary. This can be achieved through communication tools such as gRPC (google Remote Procedure Call). The communication architecture in our experiments is depicted in Fig. \ref{fig17}.

\begin{figure}
    \centering
    \includegraphics[width=0.9\linewidth]{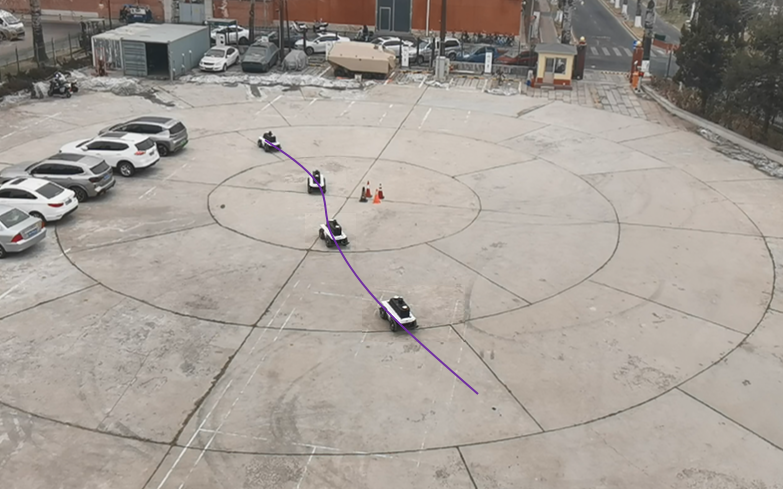}
    \caption{The trajectory of the UGV in one of the experiments.}
    \label{fig18}
\end{figure}

The experiments took place in an open square on the campus of Tsinghua University, covering an approximate area of $50m\times50m$. This setting allowed for the flexible arrangement of obstacles. During the experimental process, a global map was acquired initially and processed into a potential field, which was then converted into an uncertainty map and stored in the cloud. Following this, global path planning was carried out based on the input of the destination, and the planned path was communicated to the UGV. The onboard computer, utilizing the global path as a reference and integrating data from its own perception module, executed real-time local planning. The whole process ensured the UGV’s capability to promptly detect and avoid risks. Fig. \ref{fig18} displays the experimental setup and the UGV's trajectory in one of the trials. The global planning was completed within 0.3 seconds, while the frequency of the local planning surpassed 10Hz. The results demonstrate the feasibility and practicality of the proposed planning framework.

\section{Conclusion} \label{Sec:Conclusion} 
In this paper, we tackle the challenge of risk-aware planning for UGVs in off-road environment through a global-local planning framework. Our main idea is to integrate risk assessment methods with both global and local planning module to enhance the safety of the planning results. Within the global planning module, we employ a potential field method to assess static risk sources and introduce the coarse2fine A* algorithm for path planning. Simulation results demonstrate that, compared to the baseline algorithms, the Coarse2Fine A* algorithm not only upholds a high level of efficiency, with an average computation time of approximately 0.2 seconds, but also accomplishes a significant reduction in path uncertainty by over 30\%, thereby markedly enhancing path safety. For the local planning module, we adopt a deterministic sampling strategy, modify it to suit off-road environment and integrate a risk assessment model to emphasize the avoidance of local risks. The safety performance of the algorithm is validated through simulation results. Additionally, the algorithm's deployment on an actual UGV further confirms the practicality and effectiveness of our proposed planning framework through real-world experimentation.

In the future, our research can extend in two main areas. On one hand, the development of emergency measures is required for scenarios where the local environment is excessively harsh and no feasible local trajectory is available. On the other hand, off-road environments typically exhibit higher levels of uncertainty, demanding further consideration of the impact of environmental uncertainty in risk assessment and planning. Future work will concentrate on these two aspects.

\bibliographystyle{IEEEtran}
\bibliography{bibfile/mybibfile}

\vspace{-30pt}
\begin{IEEEbiography}[{\includegraphics[width=1in,height=1.25in,clip,keepaspectratio]{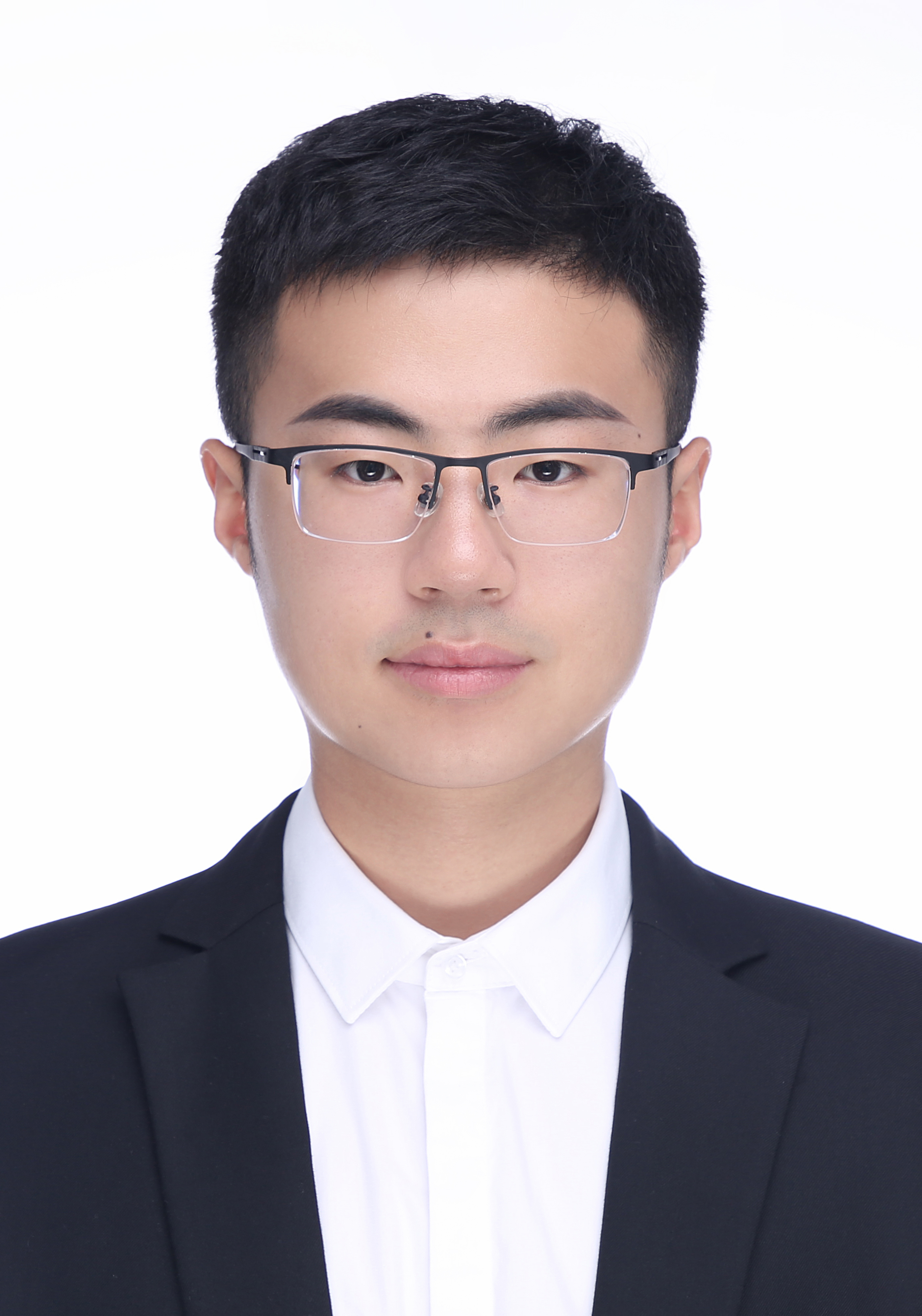}}]{Junkai~Jiang}
   received the B.E. degree from Tsinghua University, Beijing, China, in 2021, where he is currently pursuing the Ph.D. degree in mechanical engineering with the School of Vehicle and Mobility, Tsinghua University. His research interests include risk assessment, trajectory prediction, motion planning of intelligent vehicles, and multi-vehicles coordinate planning.
\end{IEEEbiography}

\vspace{-30pt}
\begin{IEEEbiography}[{\includegraphics[width=1in,height=1.25in,clip,keepaspectratio]{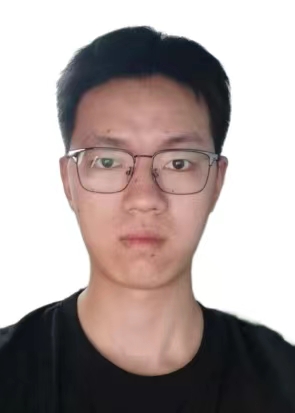}}]{Zhenhua~Hu}
   is currently pursuing the bachelor's degree in Department of Automation, Tsinghua University, Beijing. He worked as an intern at the Lab of Intelligent and Connected Vehicles, Tsinghua University, from June to December 2023. His research interests include path planning and image processing.
\end{IEEEbiography}

\vspace{-30pt}
\begin{IEEEbiography}[{\includegraphics[width=1in,height=1.25in,clip,keepaspectratio]{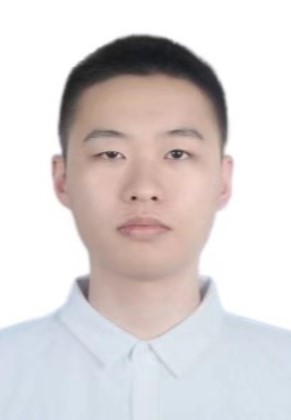}}]{Zihan~Xie}
   worked as an intern at the Lab of Intelligent and Connected Vehicles, Tsinghua University, from September to December 2023. He received the bachelor's degree in automation from the School of Mechanical, Electrical and Control Engineering at Shenzhen University, Shenzhen, in 2022. He is currently pursuing a Master's degree in Control Engineering with the School of Automation at Beijing Institute of Technology. His research interests include planning and control of intelligent vehicles.
\end{IEEEbiography}

\vspace{-30pt}
\begin{IEEEbiography}[{\includegraphics[width=1in,height=1.25in,clip,keepaspectratio]{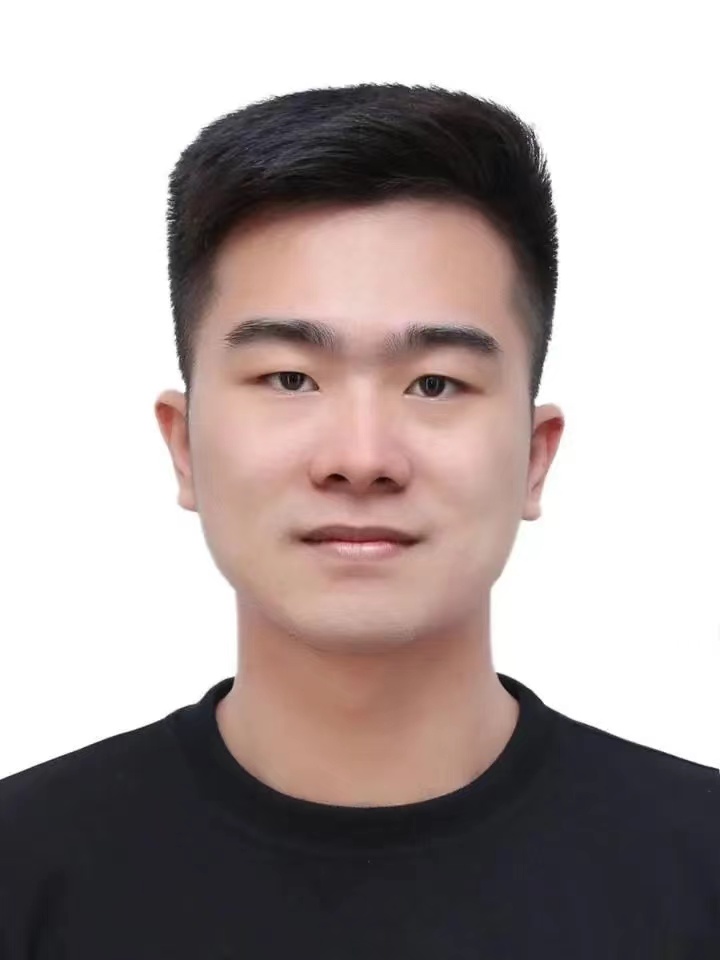}}]{Changlong~Hao}
   is currently a research intern at the Lab of Intelligent and Connected Vehicles, Tsinghua University, Beijing, China. He’s also pursuing his Master’s degree with the School of Automation, Beijing Institute of Technology, Beijing, China. His research interests include decision making, planning and control of intelligent vehicles.
\end{IEEEbiography}

\vspace{-30pt}
\begin{IEEEbiography}[{\includegraphics[width=1in,height=1.25in,clip,keepaspectratio]{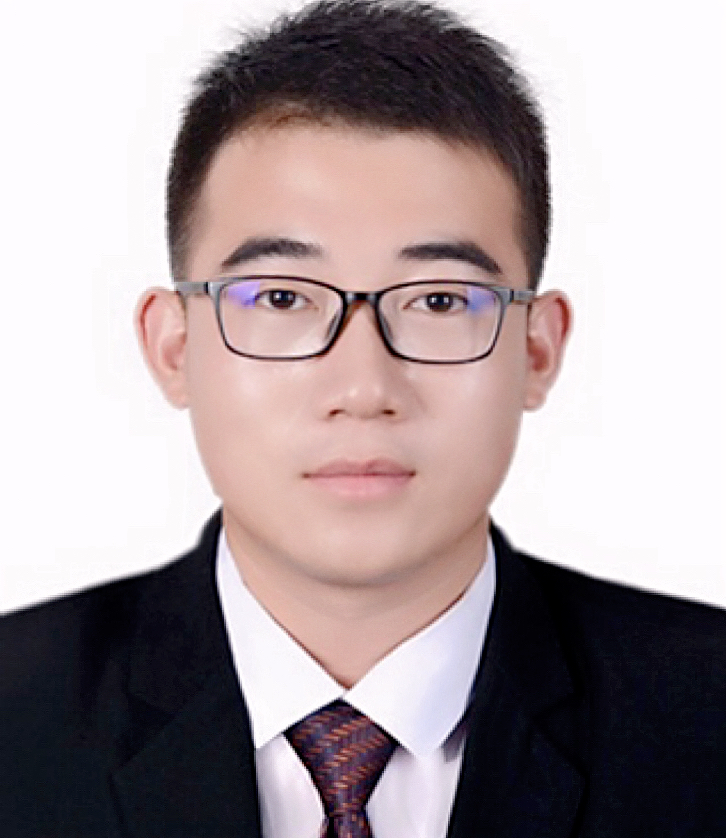}}]{Hongyu~Liu}
   worked as an intern at the Lab of Intelligent and Connected Vehicles, Tsinghua University,from September to December 2023. He received a Master's degree in Robotics and Artificial Intelligence from the University of Glasgow, UK, in January 2024. His main research interests include planning and control of intelligent vehicles.
\end{IEEEbiography}

\vspace{-30pt}
\begin{IEEEbiography}[{\includegraphics[width=1in,height=1.25in,clip,keepaspectratio]{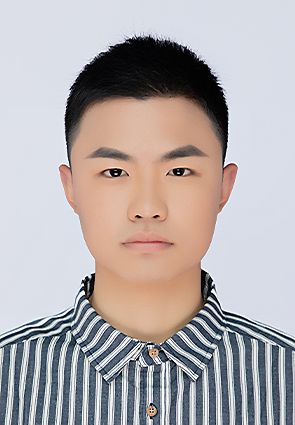}}]{Wenliang~Xu}
   worked as an intern at the lab of Intelligent and Connected Vehicles, Tsinghua University from September to December 2023. He received the bachelor's degree in Vehicle from the School of Automotive Engineering at Wuhan University of Technology, Wuhan, in 2022. He is currently pursuing a Master's degree in Vehicle Engineering with the School of Mechanical Engineering at Beijing Institute of Technology, Beijing, China. His research interests include decision making, planning and control of intelligent vehicles.
\end{IEEEbiography}

\vspace{-30pt}
\begin{IEEEbiography}[{\includegraphics[width=1in,height=1.25in,clip,keepaspectratio]{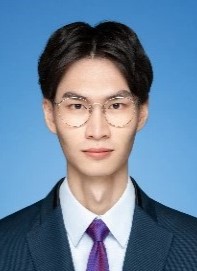}}]{Yuning~Wang}
   received the bachelor’s degree in automotive engineering from School of Vehicle and Mobility, Tsinghua University, Beijing, China, in 2020. He is currently pursuing the Ph.D. degree in mechanical engineering with School of Vehicle and Mobility, Tsinghua University, Beijing, China. His research centered on scene understanding, decision-making and planning, and driving evaluation of intelligent vehicles.
\end{IEEEbiography}

\vspace{-30pt}
\begin{IEEEbiography}[{\includegraphics[width=1in,height=1.25in,clip,keepaspectratio]{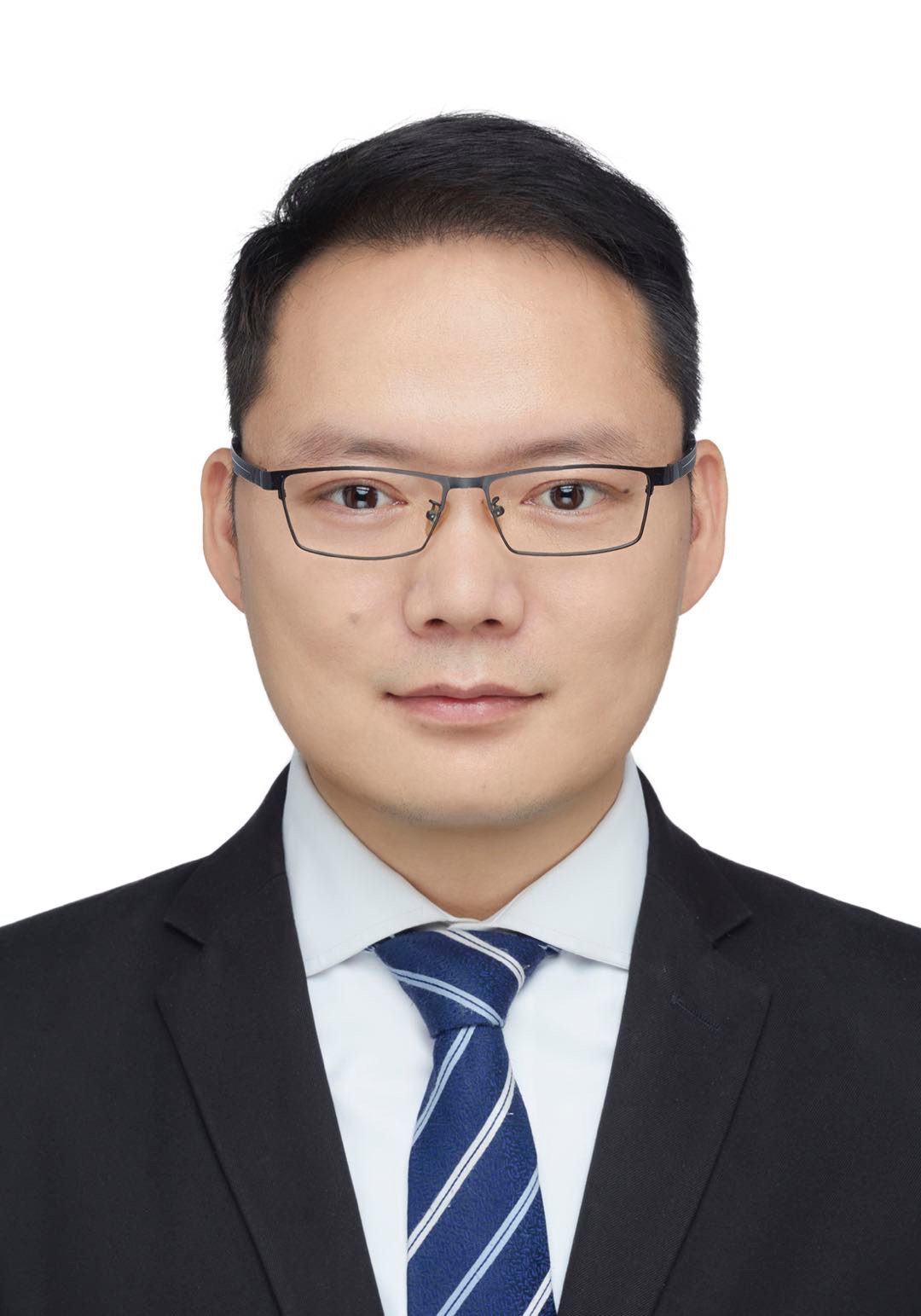}}]{Lei~He}
   received his B.S. in Beijing University of Aeronautics and Astronautics, China, in 2013, and the Ph.D. in the National Laboratory of Pattern Recognition, Chinese Academy of Sciences, in 2018. From then to 2021, Dr. He served as a postdoctoral fellow in the Department of Automation, Tsinghua University, Beijing, China. He worked as the research leader of the Autonomous Driving algorithm at Baidu and NIO from 2018 to 2023. He is a Research Scientist in automotive engineering with Tsinghua University. His research interests include Perception, SLAM, Planning, and Control.
\end{IEEEbiography}

\vspace{-30pt}
\begin{IEEEbiography}[{\includegraphics[width=1in,height=1.25in,clip,keepaspectratio]{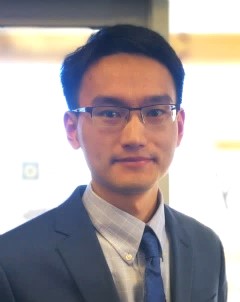}}]{Shaobing~Xu}
   received his Ph.D. degree in Mechanical Engineering from Tsinghua University, Beijing, China, in 2016. He is currently an assistant professor with the School of Vehicle and Mobility at Tsinghua University, Beijing, China. He was an assistant research scientist and postdoctoral researcher with the Department of Mechanical Engineering and Mcity at the University of Michigan, Ann Arbor. His research focuses on vehicle motion control, decision making, and path planning for autonomous vehicles. He was a recipient of the outstanding Ph.D. dissertation award of Tsinghua University and the Best Paper Award of AVEC’2018.
\end{IEEEbiography}

\vspace{-30pt}
\begin{IEEEbiography}[{\includegraphics[width=1in,height=1.25in,clip,keepaspectratio]{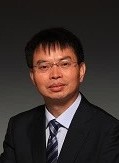}}]{Jianqiang~Wang}
   received the B.Tech. and M.S. degrees from Jilin University of Technology, Changchun, China, in 1994 and 1997, respectively, and the Ph.D. degree from Jilin University, Changchun, in 2002. He is currently a Professor and the Dean of the School of Vehicle and Mobility, Tsinghua University, Beijing, China. 

He has authored over 150 papers and is a co-inventor of over 140 patent applications. He was involved in over 10 sponsored projects. His active research interests include intelligent vehicles, driving assistance systems, and driver behavior. He was a recipient of the Best Paper Award in the 2014 IEEE Intelligent Vehicle Symposium, the Best Paper Award in the 14th ITS Asia Pacific Forum, the Best Paper Award in the 2017 IEEE Intelligent Vehicle Symposium, the Changjiang Scholar Program Professor in 2017, the Distinguished Young Scientists of NSF China in 2016, and the New Century Excellent Talents in 2008.
\end{IEEEbiography}

\end{document}